\documentclass[letterpaper]{article} % DO NOT CHANGE THIS
\usepackage{aaai2026}  % DO NOT CHANGE THIS
\usepackage{times}  % DO NOT CHANGE THIS
\usepackage{helvet}  % DO NOT CHANGE THIS
\usepackage{courier}  % DO NOT CHANGE THIS
\usepackage[hyphens]{url}  % DO NOT CHANGE THIS
\usepackage{graphicx} % DO NOT CHANGE THIS
\usepackage{natbib}  % DO NOT CHANGE THIS AND DO NOT ADD ANY OPTIONS TO IT
\usepackage{caption} % DO NOT CHANGE THIS AND DO NOT ADD ANY OPTIONS TO IT
\usepackage{algorithm}
\usepackage{algorithmic}

\usepackage{newfloat}
\usepackage{listings}
\DeclareCaptionStyle{ruled}{labelfont=normalfont,labelsep=colon,strut=off} % DO NOT CHANGE THIS
\floatstyle{ruled}
\newfloat{listing}{tb}{lst}{}
\floatname{listing}{Listing}
\usepackage{amsmath, amssymb}
\usepackage{siunitx}
\usepackage{booktabs}
\usepackage{multirow}
\usepackage[table]{xcolor}
\usepackage{microtype}

\usepackage{tikz}

\title{DOS: Distilling Observable Softmaps of Zipfian Prototypes for Self-Supervised Point Representation}
\author{
    Mohamed Abdelsamad\textsuperscript{\rm 1,2}, Michael Ulrich\textsuperscript{\rm 1}, Bin Yang\textsuperscript{\rm 1}, Miao Zhang\textsuperscript{\rm 1},\\Yakov Miron\textsuperscript{\rm 1}, Abhinav Valada\textsuperscript{\rm 2}\\
}
\affiliations{
    \textsuperscript{\rm 1}Bosch Center for Artificial Intelligence\\
    \textsuperscript{\rm 2}University of Freiburg\\
}

\begin{document}

\maketitle

\begin{abstract}
Recent advances in self-supervised learning (SSL) have shown tremendous potential for learning 3D point cloud representations without human annotations. However, SSL for 3D point clouds still faces critical challenges due to irregular geometry, shortcut-prone reconstruction, and unbalanced semantics distribution. In this work, we propose \textit{DOS} (Distilling Observable Softmaps), a novel SSL framework that self-distills semantic relevance softmaps only at observable (unmasked) points. This strategy prevents information leakage from masked regions and provides richer supervision than discrete token-to-prototype assignments. 
To address the challenge of unbalanced semantics in an unsupervised setting, we introduce Zipfian prototypes and incorporate them using a modified Sinkhorn-Knopp algorithm, \textit{Zipf-Sinkhorn}, which enforces a power-law prior over prototype usage and modulates the sharpness of the target softmap during training. 
DOS outperforms current state-of-the-art methods on semantic segmentation and 3D object detection across multiple benchmarks, including nuScenes, Waymo, SemanticKITTI, ScanNet, and ScanNet200, without relying on extra data or annotations. Our results demonstrate that observable-point softmaps distillation offers a scalable and effective paradigm for learning robust 3D representations.
\end{abstract}

% \begin{links}
%     \link{Code}{https://aaai.org/example/code}
%     \link{Datasets}{https://aaai.org/example/datasets}
%     \link{Extended version}{https://aaai.org/example/extended-version}
% \end{links}

\begin{figure}[!ht]
  \centering
  \begin{tikzpicture}
    \node[inner sep=0] (img) {
      \includegraphics[
        trim={3.5cm 16.4cm 3.5cm 5.0cm},
        clip,
        width=0.99\linewidth
      ]{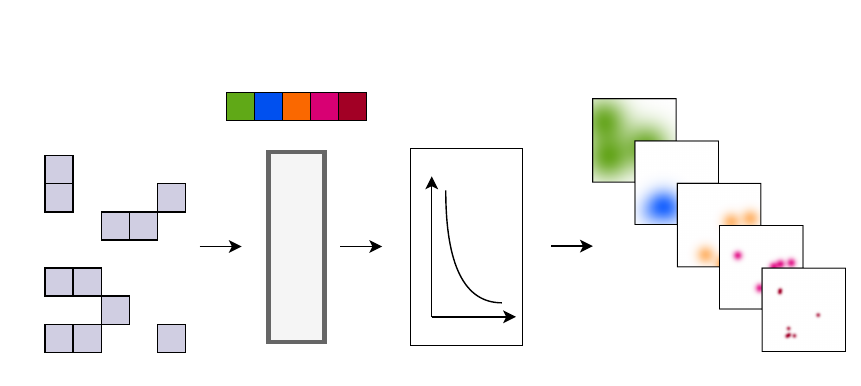}
    };

    % Coordinates relative to the picture:
    \node[anchor=north west,align=center] at ([xshift=1mm,yshift=0mm]img.north west) {
       \small Observable\\\small Point Cloud
    };
    \node[anchor=north west,align=center] at ([xshift=20mm,yshift=0mm]img.north west) {
       \small Prototypes\\\small $\mathcal{Q}$
    };
    \node[anchor=north west,align=center] at ([xshift=35mm,yshift=0mm]img.north west) {
      \small Long-tail\\\small Regularization
    };
    \node[anchor=north west,align=center] at ([xshift=63mm,yshift=0mm]img.north west) {
      \small Softmaps
    };
    \node[anchor=center,rotate=90] at ([xshift=-13mm,yshift=-5mm]img.center) {
      \small Similarity
    };
  \end{tikzpicture}
  \hfill
  \includegraphics[width=0.45\linewidth]{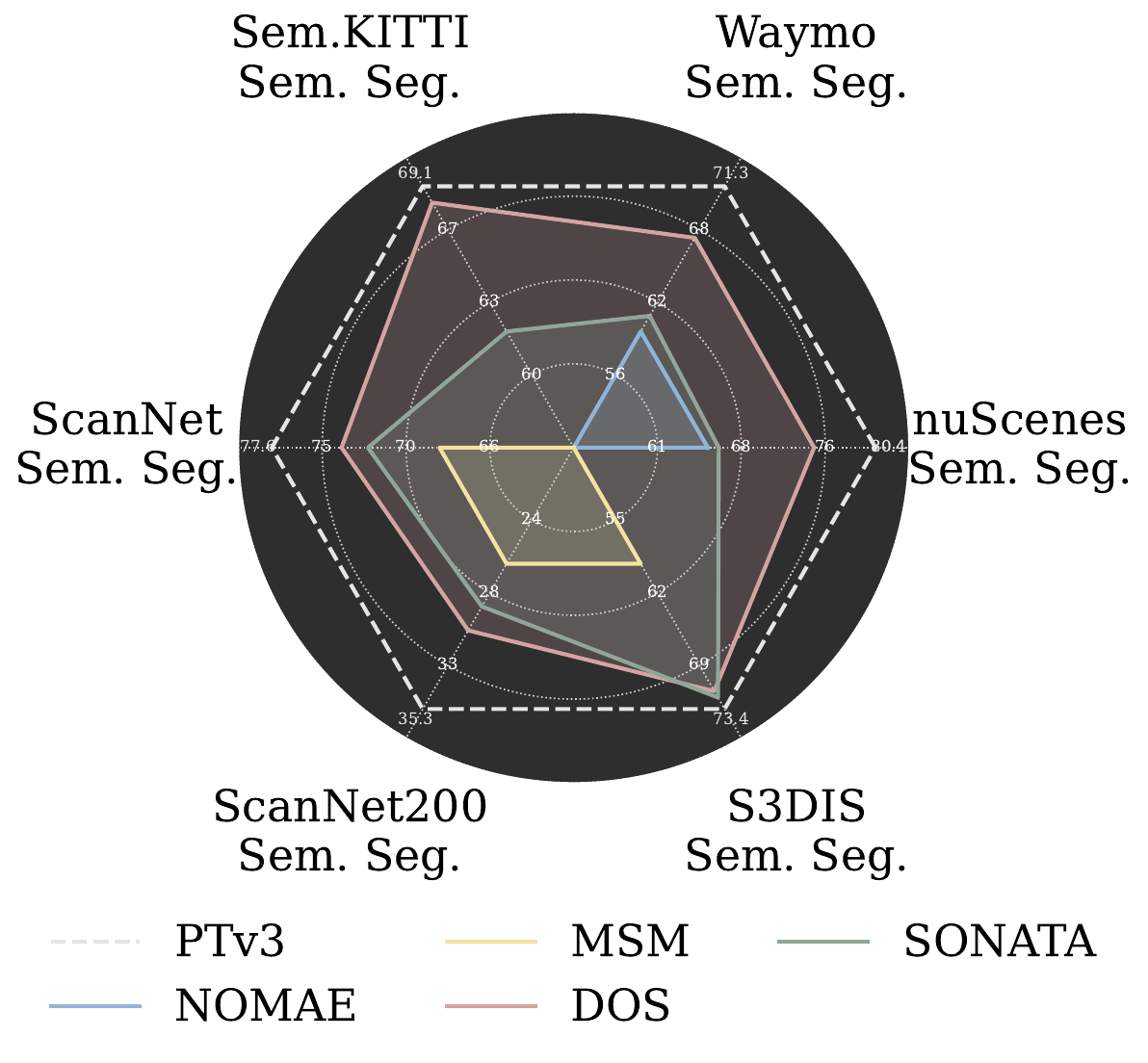}
  \includegraphics[trim={4cm 0 0cm 0},clip,width=0.54\linewidth]{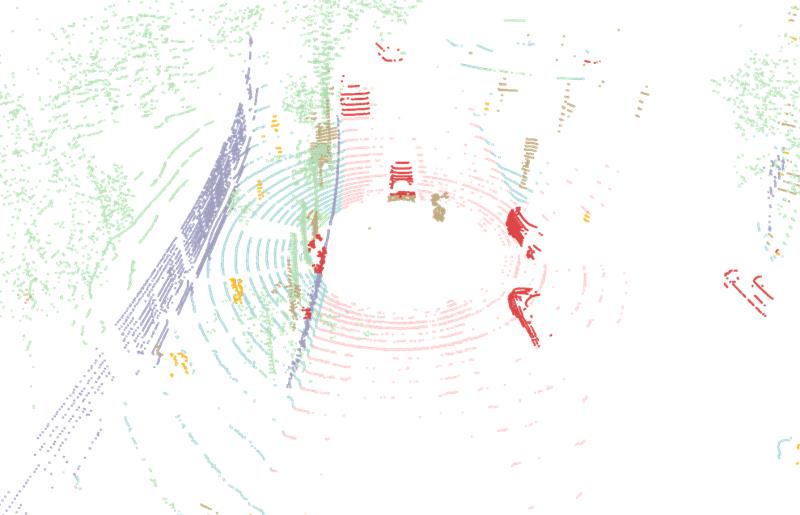}

  \caption{\textit{Overview of DOS.} \textit{Top:} DOS learns label-free 3D representations by distilling semantic softmaps at visible points, with prototype similarity and long-tail regularization. \textit{Left:} DOS under linear probing outperforms prior SSL methods on six 3D segmentation benchmarks, learning strong representation in both indoor and outdoor scenes. \textit{Right:} Provided with five samples only, a frozen DOS successfully segments the objects in a completely new domain.
  \looseness=-1}
  \label{fig:teaser}
\end{figure}
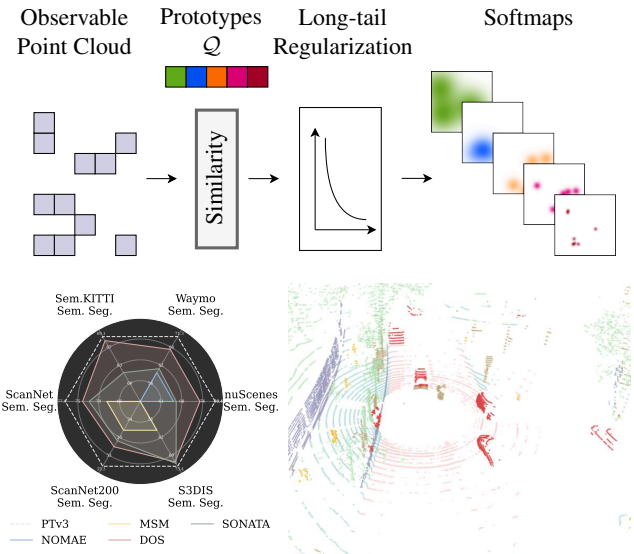

\section{Introduction}

SSL has emerged as a compelling alternative to supervised pretraining for 3D point cloud representation learning, enabling large-scale training without the need for expensive, labor-intensive annotations. While recent methods~\cite{sonata,nomae,ditr,msm,lang2024point} have demonstrated promising results across 3D perception tasks, several fundamental challenges persist. The irregular and unstructured nature of point clouds, their susceptibility to shortcut learning, and the inherently long-tailed distribution of semantic categories all contribute to a persistent performance gap between supervised and self-supervised representations on downstream tasks.\looseness=-1

Most existing SSL methods for 3D point clouds follow either reconstruction or distillation paradigms. Reconstruction-based approaches~\cite{nomae,geomae} often rely on low-level geometry, encouraging shortcut learning rather than semantic understanding. Distillation methods that leverage 2D vision models~\cite{ditr} inherit architectural biases and struggle to exploit 3D-specific structure. Recent masked self-distillation methods~\cite{sonata,msm} aim to learn semantic representations, but suffer from \textit{positional leakage}, where masked regions inadvertently influence supervision. Clustering-based objectives introduce additional instability, especially in ambiguous or imbalanced regions, while feature regression treats all points equally, ignoring semantic salience.\looseness=-1

To address these limitations, we propose \textit{DOS} (Distilling Observable Softmaps), a novel SSL framework that combines \textit{observable self-distillation} with \textit{semantic softmaps}. Observable self-distillation supervises only visible (unmasked) points, mitigating leakage and ensuring features are deduced solely from observable geometry. Softmaps encode prototype activations across space, encouraging point-wise semantic competition and providing richer gradients than clustering or feature regression. This formulation allows the model to learn nuanced spatial variation and better capture rare semantic concepts.
Despite improved stability, softmaps can suffer from degraded semantic diversity. While centering strategies help regularize and mitigate this, they also enforce uniformity, which misaligns with the naturally long-tailed distribution of 3D semantics. To address this, we introduce \textit{Zipf-Sinkhorn}, a frequency-aware assignment strategy that replaces the uniform prior with a Zipfian distribution, better reflecting real-world semantics and yielding more robust, diverse representations.\looseness=-1

We evaluate DOS on semantic segmentation and object detection tasks across six diverse benchmarks, covering both indoor and outdoor domains, and demonstrate that it consistently outperforms prior methods, with strong few-shot and cross-domain transfer capabilities. To summarize, our main contributions are as follows:
\begin{itemize}
    \item We introduce \textit{observable self-distillation}, which eliminates positional leakage by restricting supervision to unmasked points.
    \item We propose \textit{semantic softmaps} as a distillation target, promoting inter-point competition and rich spatial reasoning.
    \item We develop \textit{Zipf-Sinkhorn}, a prototype assignment scheme aligned with the long-tailed nature of real-world semantics.
    \item We release a general-purpose LiDAR backbone pretrained across multiple datasets, enabling strong transfer across 3D domains, and systematically study its generalization capabilities through cross-domain and few-shot evaluations.

\end{itemize}

\noindent
DOS achieves state-of-the-art results across five 3D benchmarks—reaching up to \textit{95\% of supervised performance} under linear probing, \textit{surpassing supervised baselines} under full fine-tuning, and significantly improving 3D object detection, especially in low-label settings. It transfers effectively to unseen domains without adaptation. Additionally we conduct a thorough ablations that confirm the effectiveness of each proposed component.

\section{Related Work}

\subsubsection{SSL for 3D Scene Understanding.}

Self-supervised learning (SSL) has become a key technique for learning 3D representations without manual annotations, particularly for scene understanding with LiDAR and RGB-D data. However, the irregular structure of point clouds demands SSL methods distinct from those in 2D vision.
Early 3D SSL methods adopt masked modeling~\cite{pointbert,geomae}, where models reconstruct occluded regions from visible geometry. These approaches often leak positional cues through masked tokens, enabling shortcut learning that limits semantic abstraction. NOMAE~\cite{nomae} mitigates leakage by reconstructing local neighborhoods, but remains sensitive to density and susceptible to geometric shortcuts. Recent methods~\cite{sonata,msm} adopt self-distillation for semantic abstraction, yet face two key limitations: supervision still relies on masked tokens, leaving positional leakage unresolved; and clustering-based objectives assume uniform prototype usage, hindering long-tail semantic modeling. We address both challenges in this work.

\subsubsection{Structured Relevance and Spatial Supervision.} A longstanding line of research in vision explores the use of spatially structured signals to improve representation learning. In 2D settings, saliency maps and class activation techniques~\cite{gradcam} highlight task-relevant regions to guide attention. Other approaches~\cite{wang2017learning,zhang2020revisiting,lang2023self,lang2024self} leverage spatial relevance maps to weight supervision, encouraging the network to focus on semantically informative areas. Relevance modeling remains underexplored in self-supervised 3D learning despite its success in supervised settings.\looseness=-1

\subsubsection{Long-Tail Distribution in 3D Data.}

Real-world 3D scenes exhibit imbalanced semantics, with a few frequent and many rare categories~\cite{ccanakcci2025label,hindel2025label}. While long-tail learning has been studied in supervised settings~\cite{focal, tao2020few, mohan2025open}, it remains underexplored in self-supervised learning, where most methods assume uniform class or prototype distributions. SwAV~\cite{swav} mitigates collapse by enforcing balanced prototype usage via the Sinkhorn algorithm, but this clashes with the natural frequency imbalance. Zipf’s law~\cite{powerlaw}, which captures power-law distributions common in vision and language, offers a more realistic semantic prior. In this work, we explicitly integrate such a prior into a self-supervised 3D framework.

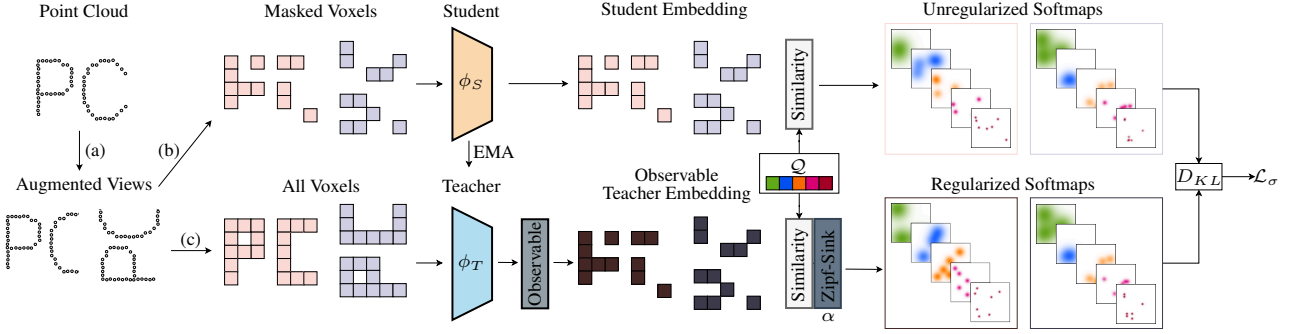
\begin{figure*}[t]
  \raggedright
  \begin{tikzpicture}
    \node[inner sep=0] (img) {
      \includegraphics[trim=0.7cm 23.5cm 0.5cm 0.5cm, clip,width=0.95\textwidth]{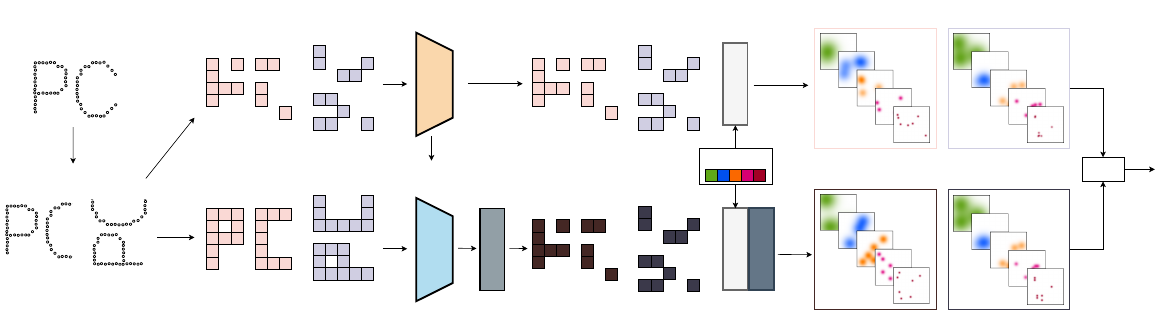}
    };

    % Coordinates relative to the picture:
    \node[anchor=center,align=center] at ([xshift=11.5mm,yshift=-2.5mm]img.north west) {
        \scriptsize Point Cloud
    };
    \node[anchor=center,align=center] at ([xshift=43mm,yshift=-2.5mm]img.north west) {
        \scriptsize Masked Voxels
    };
    \node[anchor=center,align=center] at ([xshift=63mm,yshift=-2.5mm]img.north west) {
        \scriptsize Student
    };
    \node[anchor=center,align=center] at ([xshift=90mm,yshift=-2.5mm]img.north west) {
        \scriptsize Student Embedding
    };
    \node[anchor=north west,align=center,rotate=90] at ([xshift=104mm,yshift=-18.5mm]img.north west) {
       \scriptsize Similarity
    };
    \node[anchor=center,align=center] at ([xshift=135mm,yshift=-2.5mm]img.north west) {
        \scriptsize Unregularized Softmaps
    };
    \node[anchor=center,align=center] at ([xshift=11.5mm,yshift=23mm]img.south west) {
        \scriptsize Augmented Views
    };
    \node[anchor=center,align=center] at ([xshift=43mm,yshift=23mm]img.south west) {
        \scriptsize All Voxels
    };
    \node[anchor=center,align=center] at ([xshift=63mm,yshift=23mm]img.south west) {
        \scriptsize Teacher
    };
    \node[anchor=center,align=center,font=\fontsize{7}{7}\selectfont] at ([xshift=90mm,yshift=23mm]img.south west) {
        Observable\\
        Teacher Embedding
    };
    \node[anchor=north west,align=center,rotate=90] at ([xshift=104mm,yshift=-42.5mm]img.north west) {
       \scriptsize Similarity
    };
    \node[anchor=north west,align=center,rotate=90] at ([xshift=108mm,yshift=-42.5mm]img.north west) {
       \scriptsize Zipf-Sink
    };
    \node[anchor=north west,align=center] at ([xshift=108mm,yshift=-41.5mm]img.north west) {
       \scriptsize $\alpha$
    };
    \node[anchor=north west,align=center,rotate=90] at ([xshift=69mm,yshift=-43mm]img.north west) {
       \scriptsize Observable
    };

    \node[anchor=center,align=center] at ([xshift=135mm,yshift=23mm]img.south west) {
        \scriptsize Regularized Softmaps
    };
    \node[anchor=center,align=center] at ([xshift=63mm,yshift=-12mm]img.north west) {
        \scriptsize $\phi_S$
    };
    \node[anchor=center,align=center] at ([xshift=63mm,yshift=-36mm]img.north west) {
        \scriptsize $\phi_T$
    };
    \node[anchor=center,align=center] at ([xshift=159.8mm,yshift=24.1mm]img.south west) {
        \scriptsize $D_{KL}$
    };
    \node[anchor=center,align=center] at ([xshift=168.7mm,yshift=24.1mm]img.south west) {
        \scriptsize $\mathcal{L}_\sigma$
    };
    \node[anchor=center,align=center,font=\scriptsize] at ([xshift=106.4mm,yshift=25.5mm]img.south west) {
        $\mathcal{Q}$
    };
    \node[anchor=center,align=center,font=\scriptsize] at ([xshift=66mm,yshift=27.5mm]img.south west) {
        EMA
    };
    \node[anchor=center,align=center,font=\scriptsize] at ([xshift=13mm,yshift=27.5mm]img.south west) {
        (a)
    };
    \node[anchor=center,align=center,font=\scriptsize] at ([xshift=22.8mm,yshift=27.5mm]img.south west) {
        (b)
    };
    \node[anchor=center,align=center,font=\scriptsize] at ([xshift=25.7mm,yshift=15.7mm]img.south west) {
        (c)
    };

  \end{tikzpicture}
  \centering
    \caption{
    Overview of the proposed \textit{DOS} framework. (a)~A point cloud is augmented into two views. (b)~The views are masked, voxelize and passed through the student; (c)~the corresponding unmasked views go through the teacher, and outputs are filtered to retain only observable voxels. For clarity, only same-view supervision is shown. Both student and teacher compute similarity to prototypes $\mathcal{Q}$, followed by spatial normalization into softmaps. Teacher softmaps are regularized with Zipf-Sinkhorn to reflect long-tail semantics. The student learns to match these via KL divergence.
    }

  \label{fig:framework}
\end{figure*}

\section{Method: Distilling Observable Softmaps}

In this section, we begin by outlining the general student-teacher framework shared with recent masked distillation methods. We then introduce our key contributions: \textit{observable}-point distillation to avoid masked token leakage, \textit{Softmap} loss for a richer learning signal, and \textit{Zipf-Sinkhorn} regularized softmaps to handle long-tail semantics.

\subsection{General Framework}

We adopt a standard masked self-distillation framework based on a student-teacher architecture, commonly used in recent 3D SSL approaches~\cite{sonata, msm}. As illustrated in Figure~\ref{fig:framework}, a point cloud \( \mathcal{P} = \{(\mathbf{x}_i, \mathbf{f}_i)\}_{i=1}^{N} \), where \( \mathbf{x}_i \in \mathbb{R}^3 \) denotes the spatial coordinates and \( \mathbf{f}_i \in \mathbb{R}^d \) the feature of point \( i \), is randomly cropped and duplicated into two different views, \( \mathcal{P}^{(1)} \) and $\mathcal{P}^{(2)}$. Each view undergoes separate spatial and photometric augmentations while preserving the original point coordinates to enable cross-view correspondence.
For each augmented view, we apply a random point-wise mask by selecting a visible indices subset \( \mathcal{I}_v \subset \{1, \ldots, N\} \), yielding the visible input \( \mathcal{P}_v = \{(\mathbf{x}_i, \mathbf{f}_i)\}_{i \in \mathcal{I}_v} \). The student network processes only \( \mathcal{P}_v \), while the teacher processes the full point cloud \( \mathcal{P} \). Both teacher and student share the same network architecture, and the teacher’s weights are updated via an exponential moving average (EMA) of the student parameters. 

This architectural design provides a modular setup for masked distillation and is shared with recent methods.However, it also inherits several limitations: reliance on positional cues from masked tokens, supervision focusing on point similarity, and unawareness of long-tail distributions.
We address these issues with three core changes: (1) we distill only from observable (unmasked) points to avoid information leakage, (2) we introduce \textit{Softmap}, a soft relational objective that distills prototype-specific softmaps across points, and (3) regularizing with \textit{Zipf-Sinkhorn}, an optimal transport method aligned with real-world frequency distributions.

\subsection{Observable Point Distillation}

Standard masked distillation methods supervise the student at masked tokens using feature regression~\cite{msm} or prototype contrastive losses~\cite{sonata}. However, recovering these masked tokens often relies on their positional embeddings to infer features, leading to information leakage and shortcut learning.
To avoid this, we discard masked tokens entirely and apply supervision only at observable (unmasked) points, the subset actually seen by the student. The teacher, in contrast, processes the full input \( \mathcal{P} \) and produces targets over the visible subset \( \mathcal{I}_v \). Importantly, since the teacher’s output reflects the full context while the student only sees partial input, the student is implicitly encouraged to reason about the missing (masked) regions. This encourages context-aware learning without access to positional cues from the masked areas.

\subsection{Softmap: Semantic Softmaps Distillation}
Most existing distillation methods, whether based on feature regression~\cite{msm, ditr} or soft clustering~\cite{sonata}, align teacher and student outputs at each point independently. While effective, such objectives overlook a key aspect of semantic understanding: the relative importance of different points for a given concept.
Instead of comparing distributions across features at fixed locations, we supervise the student based on how each prototype's activation is distributed across points. Specifically, the student learns to predict a \textit{semantic softmap}, a normalized relevance map indicating where a prototype is active. The target is the teacher's normalized activation across the visible points, reframing distillation as a distribution-matching problem across space rather than features.

Formally, let $\phi_T(\mathcal{P})$ and $\phi_S(\mathcal{P}_v)$ be the teacher and student embeddings, and $\mathcal{Q} = \{q_k\}_{k=1}^{K}$ the learnable prototype set. We compute cosine similarity between point embeddings and prototypes via:
\[
\begin{array}{l}
s^T_{ik} = \exp\left( \dfrac{ \cos(\phi_T(\mathcal{P})_i, q_k) }{ \tau_T } \right),\\
s^S_{ik} = \exp\left( \dfrac{ \cos(\phi_S(\mathcal{P}_v)_i, q_k) }{ \tau_S } \right),
\end{array}
\]
where $\tau_T, \tau_S$ are temperature parameters, $i \in \mathcal{I}_v$ indexes visible points, and $k$ indexes prototypes.To construct softmaps, these similarities are normalized across visible points:
\[
S_T(i, k) = \frac{s^{T}_{ik}}{\sum_{j \in \mathcal{I}_v} s^{T}_{jk}}, \quad S_S(i, k) = \frac{s^{S}_{ik}}{\sum_{j \in \mathcal{I}_v} s^{S}_{jk}}.
\]
This contrasts with clustering-based soft targets normalization $S^{\text{c}}_T(i, k) = {s^{T}_{ik}}/\sum_{k'=1}^{K} s^{T}_{ik'}$, where scores are normalized across prototypes for each point independently.
Next, we apply Zipf-Sinkhorn to regularize the teacher's softmaps using a long-tailed prior, yielding $\tilde{S}_T(i, k)$ (see Section~\ref{sec:zipf_sinkhorn} Zipf-Sinkhorn Regularization). The Softmap loss is defined as a KL divergence:
\[
\begin{aligned}
\mathcal{L}_\sigma(\mathcal{P}, \mathcal{P}_v, \mathcal{Q}) &= \frac{1}{K} \sum_{k=1}^{K} \mathrm{KL}(\tilde{S}_T(:,k) \| S_S(:,k)) \\
&= - \frac{1}{K} \sum_{k=1}^{K} \sum_{i \in \mathcal{I}_v} \tilde{S}_T(i, k) \log S_S(i, k)
\end{aligned}
\]

\begin{figure}[t]
  \centering
    \begin{tikzpicture}
    \node[inner sep=0] (img) {
\includegraphics[trim={0 0.6cm 0 0.4cm},clip,width=0.95\linewidth]{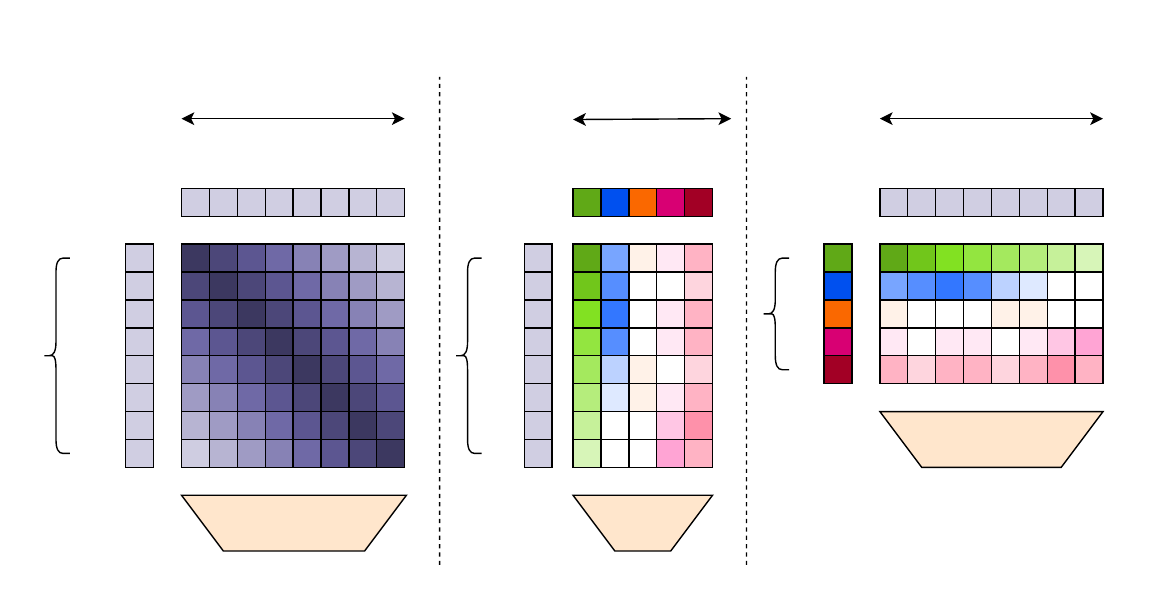}
};

    % Coordinates relative to the picture:
    \node[anchor=center,align=center,font=\fontsize{7}{7}\selectfont] at ([xshift=15mm,yshift=0.5mm]img.north west) {
       \small InfoNCE
    };
    \node[anchor=center,align=center,font=\fontsize{7}{7}\selectfont] at ([xshift=20mm,yshift=-4mm]img.north west) {
       competition\\
       Positive \& Negative
    };
    \node[anchor=center,align=center,font=\fontsize{7}{7}\selectfont] at ([xshift=40mm,yshift=0.5mm]img.north west) {
       \small Clustering
    };
    \node[anchor=center,align=center,font=\fontsize{7}{7}\selectfont] at ([xshift=43mm,yshift=-4mm]img.north west) {
       competition\\
       Positive \& Negative
    };
    \node[anchor=center,align=center,font=\fontsize{7}{7}\selectfont] at ([xshift=65mm,yshift=0.5mm]img.north west) {
       \small Softmap
    };
    \node[anchor=center,align=center,font=\fontsize{7}{7}\selectfont] at ([xshift=67mm,yshift=-4mm]img.north west) {
       competition\\
       Positive \& Negative
    };
    \node[anchor=center,align=center,font=\fontsize{7}{7}\selectfont] at ([xshift=21mm,yshift=-9.5mm]img.north west) {
       $\mathcal{P}_v$
    };
    \node[anchor=center,align=center,font=\fontsize{7}{7}\selectfont] at ([xshift=45mm,yshift=-9.5mm]img.north west) {
       $\mathcal{Q}$
    };
    \node[anchor=center,align=center,font=\fontsize{7}{7}\selectfont] at ([xshift=69mm,yshift=-9.5mm]img.north west) {
       $\mathcal{P}_v$
    };
    \node[anchor=center,align=center,font=\fontsize{7}{7}\selectfont] at ([xshift=20.5mm,yshift=-34mm]img.north west) {
       Softmax
    };
    \node[anchor=center,align=center,font=\fontsize{7}{7}\selectfont] at ([xshift=45mm,yshift=-34mm]img.north west) {
       Softmax
    };
    \node[anchor=center,align=center,font=\fontsize{7}{7}\selectfont] at ([xshift=68.7mm,yshift=-28.7mm]img.north west) {
       Softmax
    };
    \node[anchor=center,align=center,font=\fontsize{7}{7}\selectfont] at ([xshift=7mm,yshift=-13.5mm]img.north west) {
      Anchors
    };
    \node[anchor=center,align=center,font=\fontsize{7}{7}\selectfont] at ([xshift=35mm,yshift=-13.5mm]img.north west) {
      Anchors
    };
    \node[anchor=center,align=center,font=\fontsize{7}{7}\selectfont] at ([xshift=56.1mm,yshift=-13.5mm]img.north west) {
     Anchors
    };
    \node[anchor=center,align=center,font=\fontsize{7}{7}\selectfont] at ([xshift=6.5mm,yshift=-23.5mm]img.north west) {
       $\mathcal{P}_v$
    };
    \node[anchor=center,align=center,font=\fontsize{7}{7}\selectfont] at ([xshift=34.5mm,yshift=-23.5mm]img.north west) {
       $\mathcal{P}_v$
    };
    \node[anchor=center,align=center,font=\fontsize{7}{7}\selectfont] at ([xshift=55.4mm,yshift=-20mm]img.north west) {
       $\mathcal{Q}$
    };

  \end{tikzpicture}
    \caption{
        Conceptual comparison of contrastive strategies. 
        \textit{Left:} InfoNCE contrasts positive and negative pairs via pairwise similarity. 
        \textit{Middle:} Clustering aligns points to prototypes but lacks inter-point competition. 
        \textit{Right:} Softmap (DOS) distills dense soft targets and applies softmax across points, encouraging spatially structured, concept-aware representations. 
        Softmap distillation yields stronger features empirically.
    }
  \label{fig:contrastive_comparison}
\end{figure}
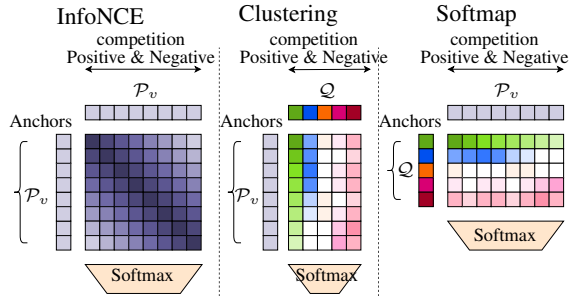

While Softmap loss does not rely on explicit positive-negative pairs, it shares a key property with InfoNCE~\cite{oord2018representation}: each prototype induces competition across points, treating them as soft positives and negatives with respect to a semantic concept. Unlike clustering-based methods that normalize over prototypes, Softmap distillation applies softmax across points (see Fig.~\ref{fig:contrastive_comparison}), shifting the objective from point-to-prototype matching to spatial reasoning. This formulation yields richer gradients, encourages spatially grounded representations, and allows even weakly activated prototypes to influence learning.

For each student view \(a \in \{1, 2\}\), we apply the Softmap loss to align the student’s predicted softmaps with Zipf-regularized teacher softmaps from both the same-view and cross-view teachers:
\[
\mathcal{L}_a = \frac{1}{2} \left( \mathcal{L}_\sigma (\mathcal{P}^{(a)}, \mathcal{P}_v^{(a)}, \mathcal{Q}) + \mathcal{L}_\sigma (\mathcal{P}^{(\bar{a})}, \mathcal{P}_v^{(a)}, \mathcal{Q}) \right),
\]
where \(\mathcal{P}_v^{(a)}\) is the masked student input, \(\mathcal{P}^{(a)}\) and \(\mathcal{P}^{(\bar{a})}\) are full inputs to the same-view and cross-view teachers, and \(\bar{a}\) denotes the opposite view index. The total training loss is
$\mathcal{L}_{\text{total}} = \mathcal{L}_1 + \mathcal{L}_2$.
This setup enforces semantic consistency between views by aligning their softmaps at observable student points. Which promotes high-level, spatially structured representations without explicit reconstruction of masked tokens.\looseness=-1

\subsection{Zipf-Sinkhorn Regularization}
\label{sec:zipf_sinkhorn}
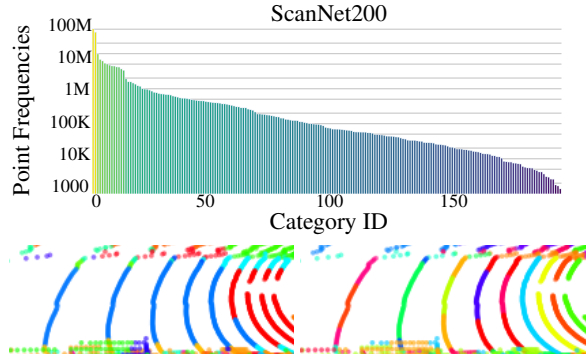
\begin{figure}[t]
  \centering
  % \begin{minipage}[t]{0.99\linewidth}
  %   \centering
  %   \includegraphics[trim={0.5cm 0cm 0cm 0.45cm},clip,width=\linewidth]{figures/scannet200.pdf}
  % \end{minipage}
  \begin{tikzpicture}
    \centering
    \node[inner sep=0] (img) {
        \includegraphics[trim={0.5cm 0cm 0cm 0.45cm},clip,width=0.8\linewidth]{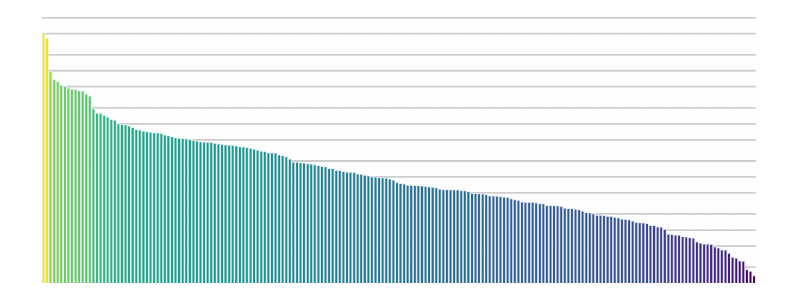}
    };
    % Coordinates relative to the picture:
    \node[anchor=center,align=center,rotate=90] at ([xshift=-8 mm,yshift=-11mm]img.north west) {
       \small Point Frequencies
    };
    \node[anchor=center,align=center,font=\fontsize{7}{7}\selectfont] at ([xshift=33mm,yshift=1.5mm]img.north west) {
       \small ScanNet200
    };
    \node[anchor=center,align=center,font=\fontsize{7}{7}\selectfont] at ([xshift=2mm,yshift=-23.5mm]img.north west) {
       0
    };
    \node[anchor=center,align=center,font=\fontsize{7}{7}\selectfont] at ([xshift=16.5mm,yshift=-23.5mm]img.north west) {
       50
    };
    \node[anchor=center,align=center,font=\fontsize{7}{7}\selectfont] at ([xshift=33mm,yshift=-23.5mm]img.north west) {
       100
    };
    \node[anchor=center,align=center,font=\fontsize{7}{7}\selectfont] at ([xshift=49.5mm,yshift=-23.5mm]img.north west) {
       150
    };
    \node[anchor=center,align=center,font=\fontsize{7}{7}\selectfont] at ([xshift=33mm,yshift=-26.5mm]img.north west) {
       \small Category ID
    };
    \node[anchor=center,align=center,font=\fontsize{7}{7}\selectfont] at ([xshift=-1.5mm,yshift=0mm]img.north west) {
       100M
    };
    \node[anchor=center,align=center,font=\fontsize{7}{7}\selectfont] at ([xshift=-1.5mm,yshift=-4.3mm]img.north west) {
       ~~10M
    };
    \node[anchor=center,align=center,font=\fontsize{7}{7}\selectfont] at ([xshift=-1.5mm,yshift=-8.6mm]img.north west) {
       ~~~1M
    };
    \node[anchor=center,align=center,font=\fontsize{7}{7}\selectfont] at ([xshift=-1.5mm,yshift=-12.9mm]img.north west) {
       100K
    };
    \node[anchor=center,align=center,font=\fontsize{7}{7}\selectfont] at ([xshift=-1.5mm,yshift=-17.2mm]img.north west) {
       ~~10K
    };
    \node[anchor=center,align=center,font=\fontsize{7}{7}\selectfont] at ([xshift=-1.5mm,yshift=-21.5mm]img.north west) {
       1000
    };
  \end{tikzpicture}
  \begin{minipage}[c]{0.45\linewidth}
    \centering
    \includegraphics[trim={3cm 8.4cm 12cm 9.9cm},clip,width=\linewidth]{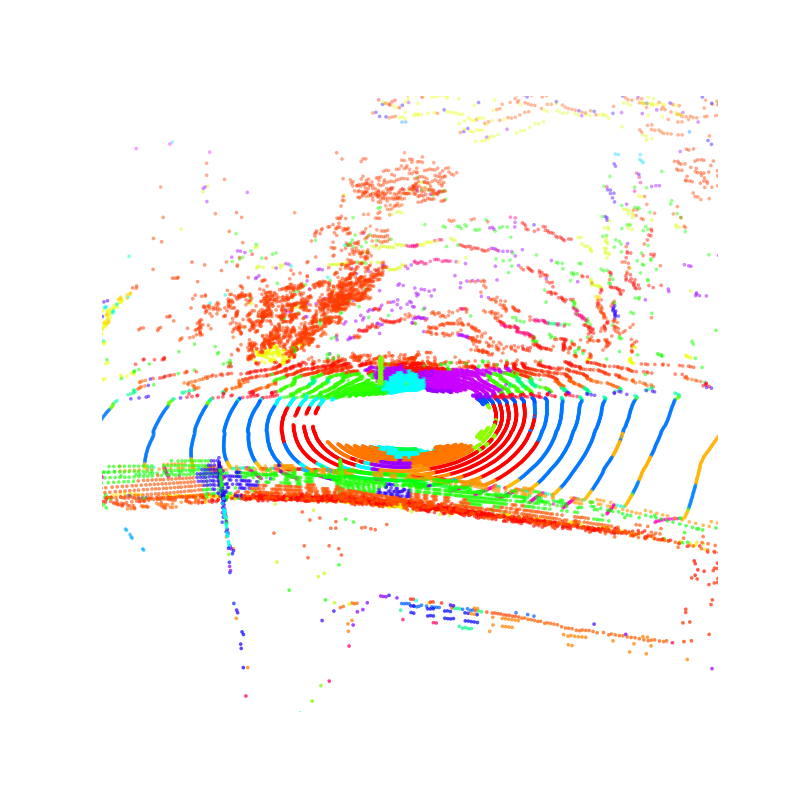}
  \end{minipage} 
  \begin{minipage}[c]{0.45\linewidth}
    \centering
    \includegraphics[trim={3cm 8.4cm 12cm 9.9cm},clip,width=\linewidth]{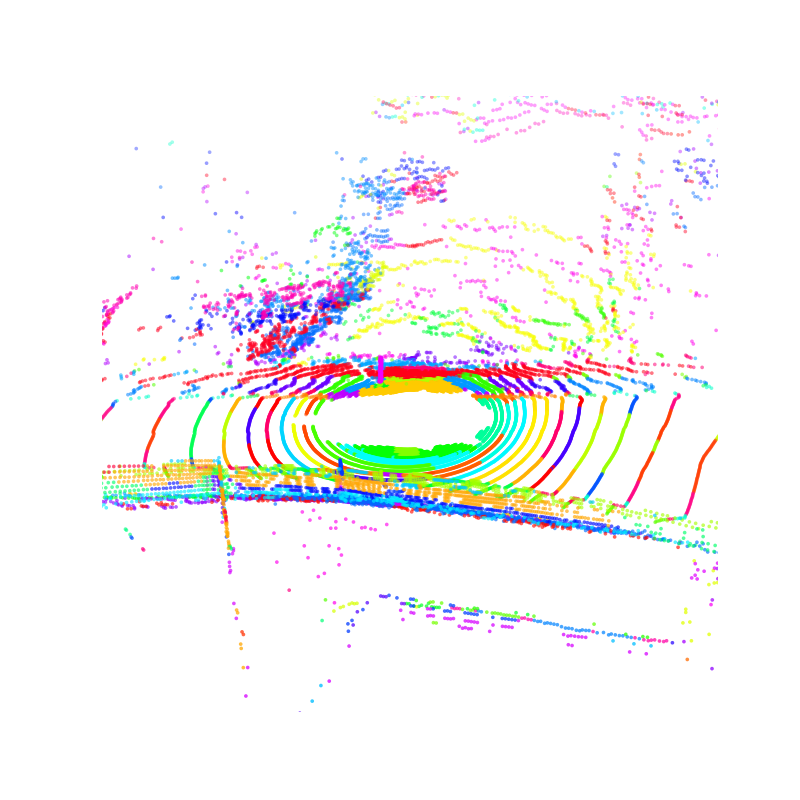}
  \end{minipage} 
\caption{
Zipfian Distribution and Prototype Assignment. 
\textit{Top:} Category frequencies in ScanNet200 follow a Zipf-like distribution, illustrating the long-tail nature of 3D semantics. 
\textit{Bottom:} Prototype activation maps after pretraining. 
\textit{Left:} Zipf prior yields consistent activations in frequent regions (e.g., road) and reduces fragmentation. 
\textit{Right:} Uniform prior enforces equal usage, leading to over-segmentation and learning biased toward low-level geometry (e.g., ego distance).
}
\label{fig:zipf_uniform_comparison}
\end{figure}

\begin{algorithm}[!h]
\caption{Zipf-Sinkhorn for Softmaps Regularization}
\label{alg:zipf_sinkhorn}
\begin{algorithmic}[1]
\REQUIRE Teacher similarities matrix $F \in \mathbb{R}^{N_v \times K}$, Zipf exponent $\alpha$, iterations $T$
\ENSURE Zipf-aware softmap $\tilde{S}_T \in \mathbb{R}^{N_v \times K}$

\vspace{1mm}
\STATE \textbf{Compute Zipf Prior:}
\STATE \hspace{2mm} $w_k \propto \frac{1}{k^\alpha}$ for $k = 1, 2, \ldots, K$, then normalize $\mathbf{w} \gets \mathbf{w} / \sum_k w_k$
\vspace{1mm}

\STATE \textbf{Normalize Similarities:}
\STATE \hspace{2mm} Normalize: $F \gets F / \sum_{i,k} F_{ik}$
\vspace{1mm}

\STATE \textbf{Iterative Normalization (Sinkhorn):}
\FOR{$t = 1$ to $T$}
    \STATE \hspace{2mm} Normalize rows: $F_{i,:} \gets F_{i,:} / \sum_k F_{i,k} \quad \forall i$
    \STATE \hspace{2mm} Normalize columns to match $\mathbf{w}$: $F_{:,k} \gets F_{:,k} \cdot \mathbf{w}_k / \sum_i F_{i,k} \quad \forall k$
\ENDFOR
\vspace{1mm}

\STATE \textbf{Column Normalize for Prototype activation:}
\STATE \hspace{2mm} $\tilde{S}_T(i, k) \gets F_{ik} / \sum_j F_{jk}$
\vspace{1mm}

\RETURN $\tilde{S}_T$
\end{algorithmic}
\end{algorithm}

 \begin{table*}[t]
    \centering
    \footnotesize
    %\vspace{-0.5em}
    \begin{tabular}{lcccccccccccc}
    \toprule
    \textbf{Method} 
    & \multicolumn{2}{c}{\textbf{nuScenes}} 
    & \multicolumn{2}{c}{\textbf{Waymo}} 
    & \multicolumn{2}{c}{\textbf{SemKITTI}} 
    & \multicolumn{2}{c}{\textbf{ScanNet}} 
    & \multicolumn{2}{c}{\textbf{ScanNet200}} 
    & \multicolumn{2}{c}{\textbf{S3DIS Area 5}} \\
    & mIoU & mAcc 
    & mIoU & mAcc 
    & mIoU & mAcc 
    & mIoU & mAcc 
    & mIoU & mAcc 
    & mIoU & mAcc \\
    \midrule
    PTv3~\shortcite{ptv3} & 80.4 & 87.2 & 71.3 & 80.5 & 69.1 & 76.1 & 77.6 & 85.0 & 35.3 & 46.0 & 73.4 & 78.9 \\
    \midrule
    MSM~\shortcite{msm} & - & - & - & - & - & - & 68.7 & - & 26.8 & - & 59.5 & - \\
    NOMAE~\shortcite{nomae} & 65.1 & 77.9 & 59.2 & 69.9 & - & - & - & - & - & - & - & -\\
    Sonata$^*$~\shortcite{sonata}  & 66.1 & 77.2 & 60.5 & 72.5 & 62.0 & 72.5 & 72.5 & 83.1 & 29.3 & 41.6 & \textbf{72.3} & 81.2\\
    \rowcolor[HTML]{E8F5E9} % light green
    DOS (ours)  & 74.1 & \textbf{84.8} & 66.1 & 77.1 & 67.5 & 78.1 & 72.8 & 83.3 & 29.1 & 41.1 & 70.6 & 79.2 \\
    \rowcolor[HTML]{E8F5E9} % light green
    DOS$^*$ (ours)  & \textbf{74.8} & 84.2 & \textbf{67.0} & \textbf{77.7} & \textbf{68.3} & \textbf{78.4} & \textbf{73.9} & \textbf{83.5} & \textbf{30.7} & \textbf{41.7} & 71.7 & \textbf{81.4} \\
    \midrule
    D-DITR~\shortcite{ditr} & 80.7 & - & 72.1 & - & 69.8 & - & 79.2 & - & \textbf{37.7} & - & 75.0 & - \\
    MSM~\shortcite{msm} & - & - & - & - & - & - & 78.5 & - & 35.7 & - & 73.2 & - \\
    NOMAE~\shortcite{nomae} & \textbf{81.8} & 87.7 &  72.3 & 82.5  & - & - & - & - & - & - & - & -\\
    Sonata$^*$~\shortcite{sonata}~ & \underline{81.7} & \textbf{87.9} & 72.9 & 81.9 & 72.6 & 77.9 & \underline{79.4} & 86.1 & 36.8 & 46.5 & \textbf{76.0} & 81.6 \\
    % \rowcolor[HTML]{E8F5E9} % light green
    % DOS (ours) (dec) & \textbf{79.9} & \textbf{86.3} & \textbf{71.7} & \textbf{80.9} & \textbf{71.1} & \textbf{77.1} & 78.3 & 85.7 & \textbf{35.5} & \textbf{46.2} & TODO & TODO \\
    \rowcolor[HTML]{E8F5E9} % light green
    DOS (ours)  & 81.5 & 87.6 & \underline{73.3} & \underline{83.8} & \underline{73.1} & \underline{81.0} & 78.7 & \underline{86.2}& 36.7 & \underline{46.6} & 74.2 & \underline{83.6} \\
    \rowcolor[HTML]{E8F5E9} % light green
    DOS$^*$ (ours)  & \textbf{81.8} & \underline{87.8}& \textbf{73.9} & \textbf{83.9} & \textbf{73.5} & \textbf{81.3} & \textbf{79.7} & \textbf{86.8}  & \underline{37.1} & \textbf{46.8} & \underline{75.1} & \textbf{83.8} \\
    \bottomrule
    % Sonata$^*$~\shortcite{sonata}~(dec) & 77.3 & 85.9 & 70.8 & 78.8 & 68.4 & 76.5 & \textbf{79.1} & \textbf{86.6} & 33.5 & 44.5 & 74.5& 80.4\\
    
    % \midrule
    % \rowcolor[HTML]{E8F5E9} % light green
    \end{tabular}
    %\vspace{-1.5em}
    \caption{Comparison of semantic segmentation performance on the validation split of multiple indoor and outdoor point cloud datasets. Methods marked with $^{*}$ use additional data. For finetuning, we \textbf{bold} the best result and \underline{underline} the second-best.}
    \label{tab:sota_semseg}
\end{table*}

Softmaps provide richer gradients and a stronger inductive bias than clustering, enabling stable training even without explicit centering. However, performance remains suboptimal due to two failure modes: (1) semantic prototype collapse, where multiple prototypes redundantly activate on the same regions; and (2) point-wise collapse, where parts of the scene are left unrepresented. To address this, we aim for prototypes that (i) activate in distinct regions and (ii) collectively provide full scene coverage. While standard Sinkhorn-Knopp promotes these goals, it additionally enforces uniform prototype usage, a rigid constraint misaligned with real-world 3D semantics, which follow a Zipfian distribution (Figure~\ref{fig:zipf_uniform_comparison}, top). As a result, uniform balancing tends to over-segment frequent structures and assign prototypes based on low-level cues like distance, reducing semantic fidelity (Figure~\ref{fig:zipf_uniform_comparison}, bottom right).

To address the mismatch between uniform prototype usage and real-world semantics, we propose \textit{Zipf-Sinkhorn}, a modified optimal transport step that incorporates a power-law prior over prototypes. Instead of enforcing a uniform marginal, we assign a Zipfian prior $\pi \in \mathbb{R}^K$, where $\pi_k \propto \frac{1}{k^\alpha}$ and $\alpha > 0$ controls the sharpness. This reflects the semantic frequency imbalance in 3D data where frequent prototypes receive broader activation, while rare ones remain sharp and selective.

Let $F \in \mathbb{R}^{N_v \times K}$ be the similarity matrix between $N_v$ visible points and $K$ prototypes. We apply an entropy-regularized Sinkhorn procedure (Algorithm~\ref{alg:zipf_sinkhorn}) to iteratively normalize $F$ such that its columns match $\pi$ and its rows remain uniform. After convergence, columns are renormalized to yield the final Zipf-aware softmap $\tilde{S}_T$.
This simple modification does not introduce additional computational cost beyond computing the Zipf prior once at initialization. However, by aligning supervision with long-tailed semantics, Zipf-Sinkhorn improves representational diversity during pretraining.

\section{Implementation Details}

We use PTv3~\cite{ptv3} with layer normalization as the default encoder. Pretraining is done on 2$\times$A100 GPUs with a batch size of 16 and completes in $\sim$20 hours, depending on dataset size. A 70\% masking ratio is applied uniformly for semantic segmentation experiments, while for object detection we use a 60\% masking ratio. We use mask block sizes of \SI{40}{\centi\meter} for indoor and \SI{1}{\meter} for outdoor datasets. We apply the pretraining supervision at a voxel size of \SI{0.08}{\meter} for indoor scenes and \SI{0.2}{\meter} for outdoor scenes. To obtain voxel features at these resolutions, we follow Sonata~\cite{sonata} and upcast features from deeper, coarser layers to the target voxel grid, then concatenate them with the features computed at that resolution. We use 1024 prototypes for all experiments. For semantic segmentation, we append a lightweight PTv3 decoder; for object detection, we use the pretrained encoder as a backbone for a CenterPoint detector~\cite{centerpoint}. During fine-tuning, we follow the standard settings used in the respective supervised counterparts. Full details are provided in the supplementary material.

\section{Experimental Evaluation}

\subsection{Semantic Segmentation Results}
\label{sec:main_results}

Table~\ref{tab:sota_semseg} presents semantic segmentation results on six 3D benchmarks, spanning diverse environments and sensors. For autonomous driving, we evaluate on {nuScenes}~\cite{nuscenes}, {Waymo}~\cite{waymo}, and {SemanticKITTI}~\cite{semantickitti}; for indoor scene understanding, we use {ScanNet}~\cite{scannet}, {ScanNet200}~\cite{scannet200}, and {S3DIS}~\cite{s3dis}. We report performance using two protocols: (i) {linear probing (LP)} with a frozen encoder and linear head, and (ii) {fine-tuning (FT)} with full encoder-decoder optimization. Additional dataset details are provided in the supplementary material.

DOS achieves consistently strong results across all benchmarks, outperforming prior SSL approaches under both LP and FT. Notably, \textit{DOS reaches up to 95\% of the supervised performance under LP} and surpasses the supervised PTv3 baseline under fine-tuning on every dataset.
Compared to NOMAE~\cite{nomae}, DOS achieves substantially higher LP performance on all outdoor datasets, indicating that self-distillation yields more semantic features than geometry-centric reconstruction. Additionally, DOS outperforms D-DITR~\cite{ditr} across nearly all metrics, despite D-DITR using multi-view images and vision-based distillation.

Compared to Sonata~\cite{sonata} and MSM~\cite{msm}, both of which use masked self-distillation, DOS offers consistent improvements across LP and FT. For example, on Waymo and nuScenes, DOS improves over Sonata by +6.5 and +8.0 mIoU (LP), and by +0.7 and +0.1 mIoU (FT). These improvements stem from DOS’s key innovations: \textit{observable self-distillation}, \textit{softmap supervision}, and \textit{Zipf-Sinkhorn regularization}, which together mitigate information leakage, reduce prototype collapse, and promote semantically diverse learning. With additional training data (denoted as DOS$^*$), the performance further improves across all settings, achieving state-of-the-art SSL results on all six benchmarks.

\subsection{Object Detection Results} 

\begin{table}
    \footnotesize
    \centering
    \bgroup
        \setlength\tabcolsep{3pt}
        \begin{tabular}{p{5.0cm} >{\centering\arraybackslash}p{1.4cm} >{\centering\arraybackslash}p{1.4cm}}
            \toprule
            \textbf{Methods} & \textbf{NDS} & \textbf{mAP} \\
            \midrule
            \multicolumn{3}{l}{\textit{20\% Labeled Frames (Data-Efficient Setting)}} \\
            ALSO~\citep{boulch2023also} &  48.2 & 41.2 \\
            GD-MAE~\citep{yang2023gd-mae} &  48.8 & 42.6 \\
            Learning from 2D~\citep{Liu2021LearningF2} & 49.2 & 48.8 \\
            UniPAD~\citep{Yang2023UniPADAU}  & 55.8 & 48.1 \\
            NOMAE~\citep{nomae} & 60.9 & 54.4 \\
            \rowcolor[HTML]{E8F5E9}
            DOS (ours) & \textbf{62.1} & \textbf{57.1} \\
            \midrule
            \multicolumn{3}{l}{\textit{Full Data (100\% Labeled)}} \\
            CenterPoint (no pre-training) & 65.4 & 57.6 \\
            \rowcolor[HTML]{E8F5E9}
            CenterPoint + DOS (ours) & \textbf{69.7} & \textbf{65.5} \\
            \bottomrule
        \end{tabular}
    \egroup
    \caption{
    Object detection on nuScenes val set under full-data (complete training set) and 20\%-label (fine-tuning with 20\% labeled frames). DOS improves performance in both regimes.
    }
    \label{tab:object_detection_combined}
\end{table}

This section presents the object detection performance on the nuScenes validation dataset. During 
Table~\ref{tab:object_detection_combined} evaluates the nuScenes detection score (NDS) and mean average precision (mAP) under two settings: full-data, where both SSL pre-training and fine-tuning use the complete training set, and a commonly adopted data-efficient regime, where fine-tuning is conducted with only 20\% of the labeled frames and relevant baselines are reported.
DOS-pretrained model improves 3D object detection on nuScenes in both low-resource and full-data regimes. In the 20\%-label setting, DOS surpasses all baselines on mAP and NDS. In the full-data regime, integrating DOS with CenterPoint significantly boosts performance, yielding a +4.3 NDS and +7.9 mAP improvement over training from scratch, highlighting the transferability of DOS to detection tasks.

\subsection{Cross-Domain Transfer Analysis}

\begin{table}
    \centering
    \footnotesize
    \bgroup
        \setlength\tabcolsep{5pt}
        \begin{tabular}{p{5em}|cccccc}
        \toprule
        \textbf{Training} 
        & \multicolumn{2}{c}{\textbf{nuScenes}} 
        & \multicolumn{2}{c}{\textbf{Waymo}} 
        & \multicolumn{2}{c}{\textbf{SemKITTI}} \\
        \textbf{dataset} 
        & mIoU & mAcc 
        & mIoU & mAcc 
        & mIoU & mAcc \\
        \midrule
        nuScenes     &  \underline{74.1} & \underline{\textbf{84.8}}  & 55.5  & 68.7  & 57.3  & 68.5  \\
        Waymo        &   66.2   &  78.1     & \underline{66.1}  & \underline{77.1}  & 64.1 & 74.8 \\
        SemKITTI     &   59.5   &   72.5    &58.5  &  71.4   & \underline{67.5}  & \underline{78.1}  \\
        \rowcolor[HTML]{E8F5E9} % light green
        \textbf{Nu-SK-Wa} &\textbf{74.8}   &84.2   & \textbf{67.0}  & \textbf{77.7}  &  \textbf{68.3} & \textbf{78.4}  \\
        \bottomrule
        \end{tabular}
    \egroup
    \caption{Cross-domain segmentation performance of DOS when pretrained on individual datasets (rows) and evaluated on different domains (columns).}
    \label{tab:multi_dataset_outdoor}
\end{table}

\begin{table}[t]
    \footnotesize
    \bgroup
        \setlength\tabcolsep{6.7pt}
        \begin{tabular}{c|c|ccc}
        \toprule
        \textbf{\# Shots} & \textbf{setting} 
        & \textbf{nuScenes} 
        & \textbf{SemKITTI} 
        & \textbf{CoLa}\\
        \midrule
        0 & supervised & 6.9 & 18.0 & 37.5 \\
        \rowcolor[HTML]{E8F5E9} % light green
        0 & DOS(ours) & 50.8  & \textbf{43.8} & 63.9 \\
        \midrule
        \rowcolor[HTML]{E8F5E9} % light green
        5 & DOS(ours) & \textbf{51.5}  & 41.9 & \textbf{85.9}
         \\
        \bottomrule
        \end{tabular}
    \egroup
    \caption{ParisLuco evaluation under zero-shot and few-shot settings. Zero-shot: model fine-tuned on nuScenes or SemanticKITTI and evaluated directly on ParisLuco. Few-shot: five annotated ParisLuco scenes used.}
    \label{tab:parisluco}
\end{table}

We evaluate cross-domain generalization by pretraining DOS on nuScenes, Waymo, or SemanticKITTI, then evaluating it on the other two. Table~\ref{tab:multi_dataset_outdoor} shows that even without joint training, DOS exhibits strong transfer. A model trained solely on Waymo surpasses all other methods, including Sonata trained on all datasets, when evaluated on nuScenes and SemanticKITTI, highlighting the robustness of the learned features.
We then train a single DOS model on all three datasets combined. This unified model performs consistently well across domains, confirming DOS’s suitability as a general-purpose 3D backbone.

\begin{figure*}
\centering
\footnotesize

% --- Header row ---
\begin{tabular*}{\textwidth}{@{\extracolsep{\fill}}cccccccc}
& \textbf{Ground Truth} & \textbf{DOS (ours)} & \textbf{PTv3~\shortcite{ptv3}} &
& \textbf{Ground Truth} & \textbf{DOS (ours)} & \textbf{PTv3~\shortcite{ptv3}} \\
\end{tabular*}

\vspace{1mm}

% --- Row 1: (a) and (b) ---
\begin{minipage}[t]{\textwidth}
\centering
\textbf{(a)}\hspace{1mm}
\includegraphics[trim={25cm 5cm 25cm 5cm},clip,width=0.15\textwidth]{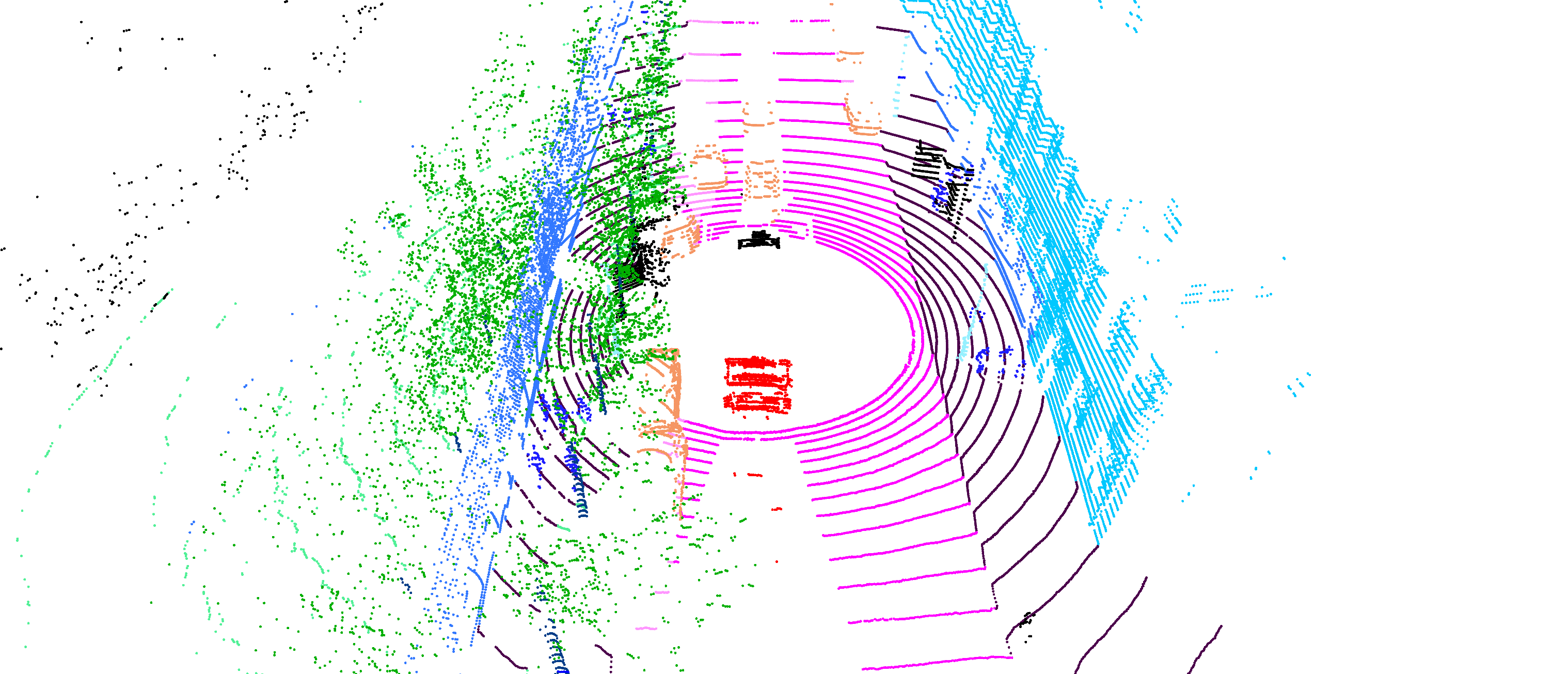}
\hfill
\includegraphics[trim={25cm 5cm 25cm 5cm},clip,width=0.15\textwidth]{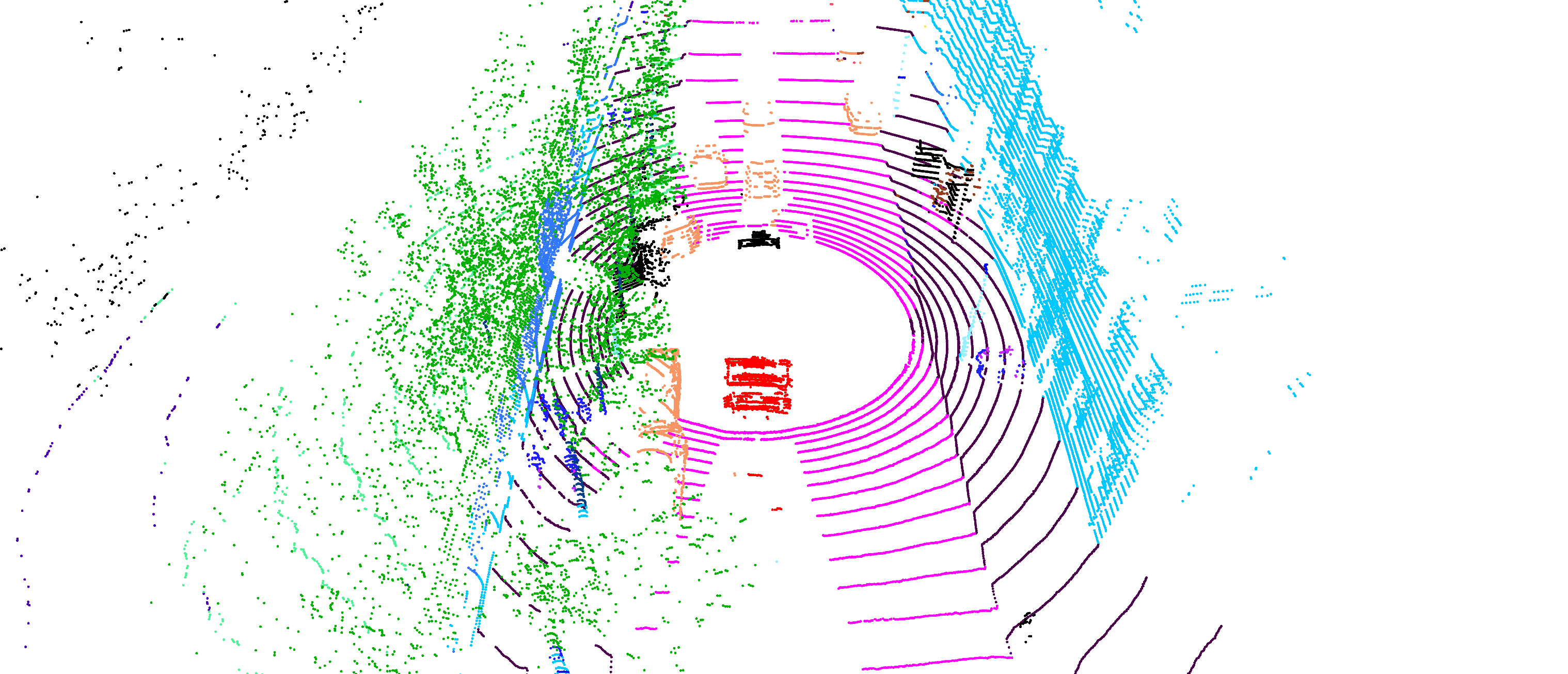}
\hfill
\includegraphics[trim={25cm 5cm 25cm 5cm},clip,width=0.15\textwidth]{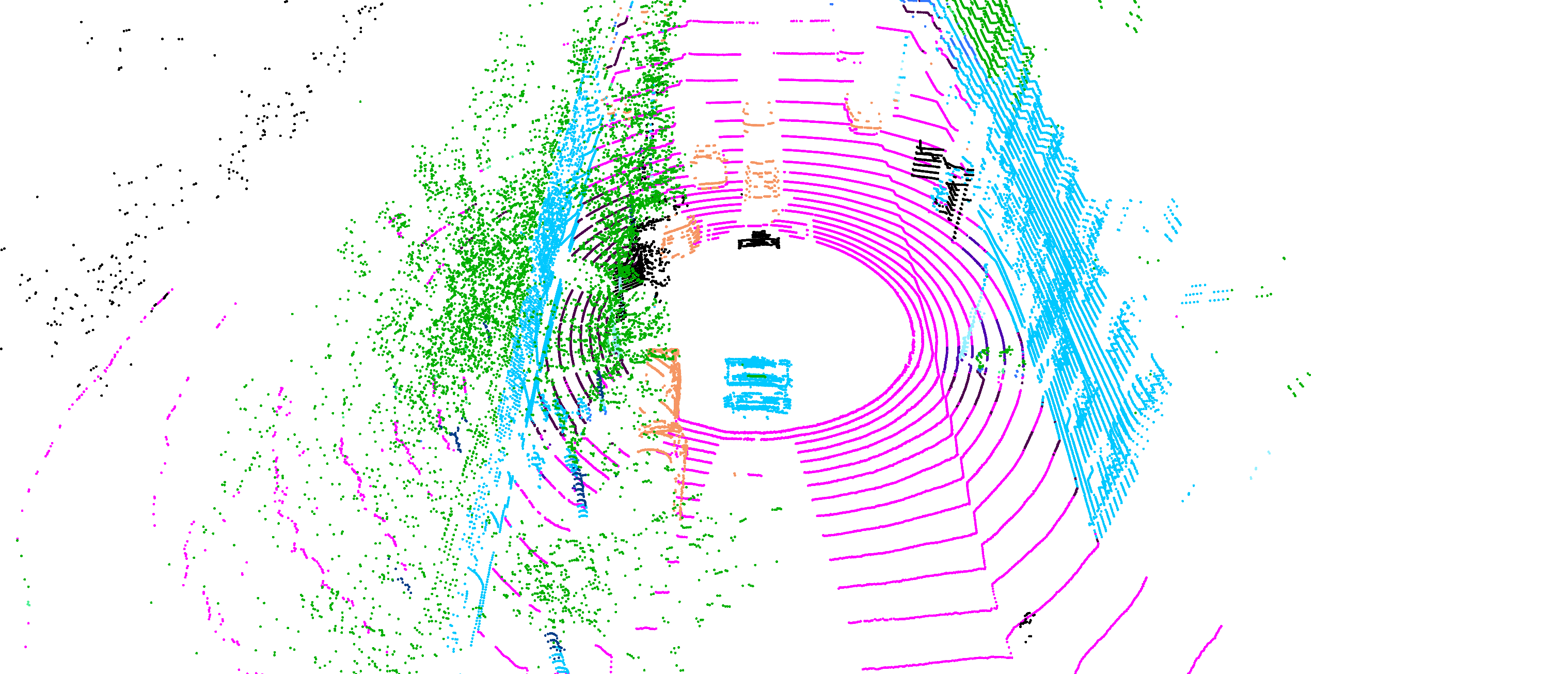}
\hspace{0.01\textwidth}
\textbf{(b)}\hspace{1mm}
\includegraphics[trim={25cm 5cm 25cm 5cm},clip,width=0.15\textwidth]{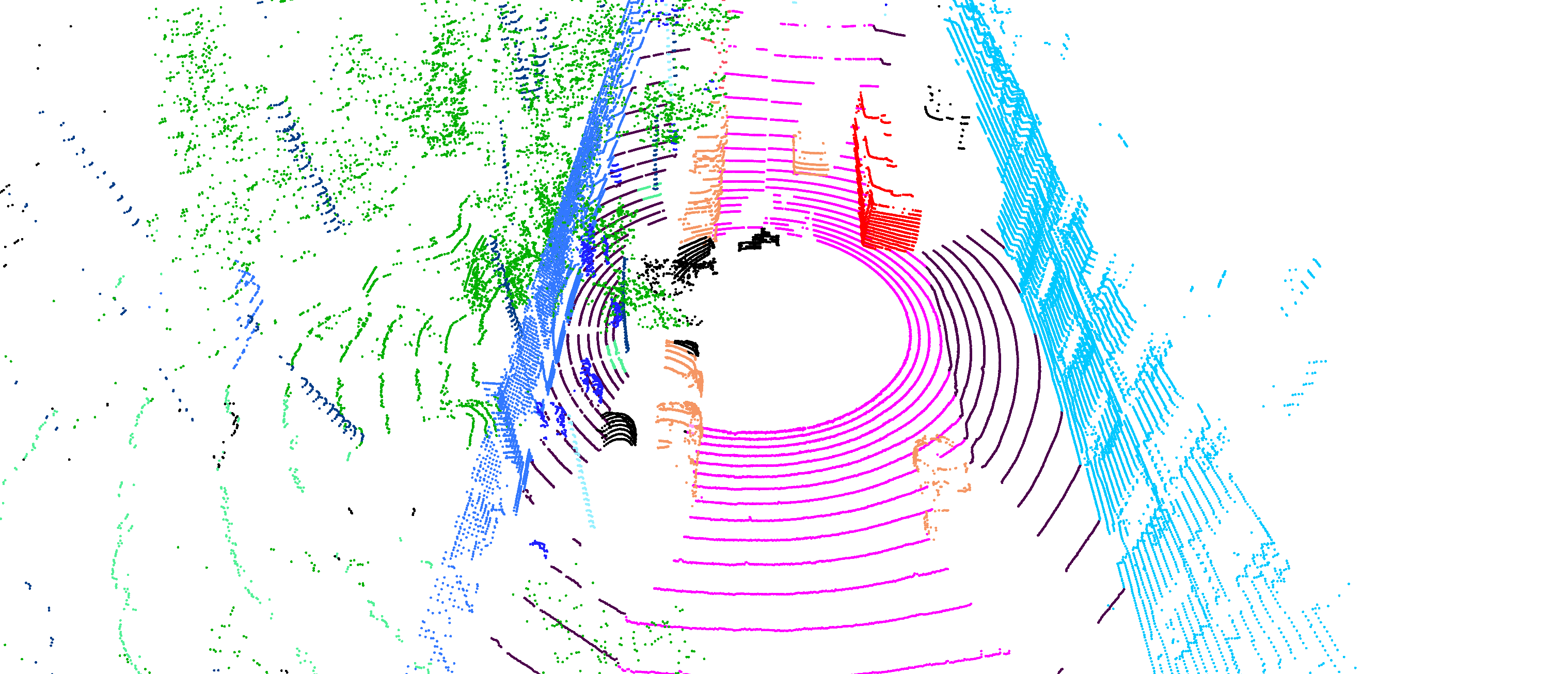}
\hfill
\includegraphics[trim={25cm 5cm 25cm 5cm},clip,width=0.15\textwidth]{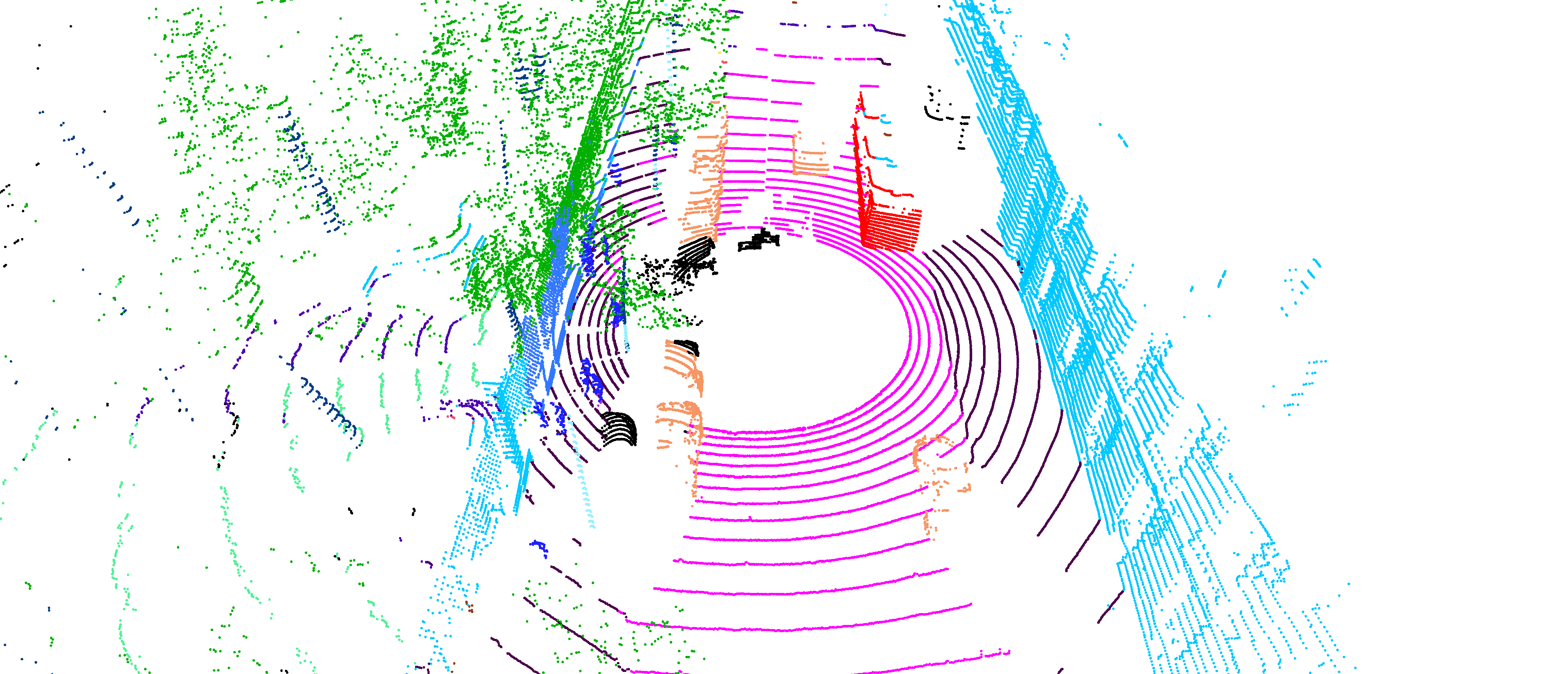}
\hfill
\includegraphics[trim={25cm 5cm 25cm 5cm},clip,width=0.15\textwidth]{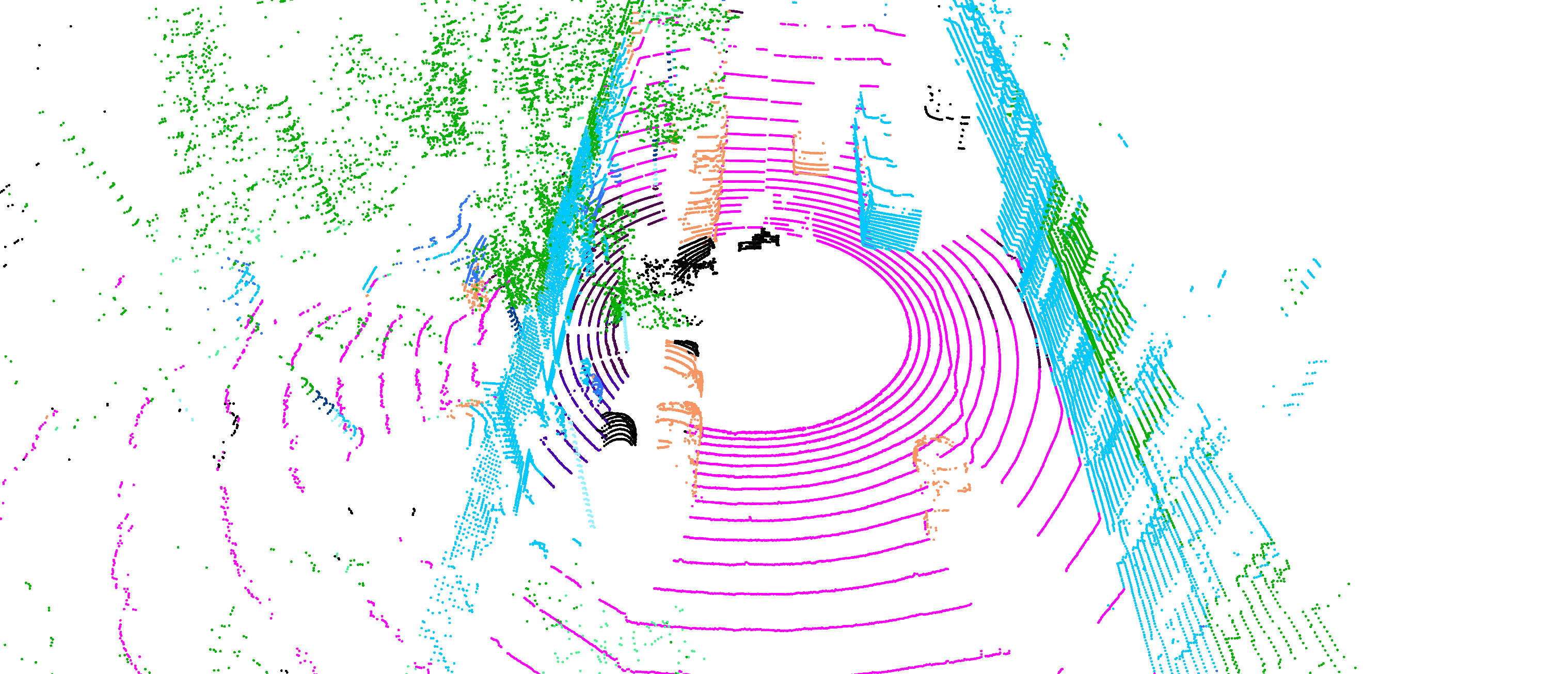}
\end{minipage}

\vspace{1mm}

% --- Row 2: (c) and (d) ---
\begin{minipage}[t]{\textwidth}
\centering
\textbf{(c)}\hspace{1mm}
\includegraphics[trim={25cm 5cm 25cm 5cm},clip,width=0.15\textwidth]{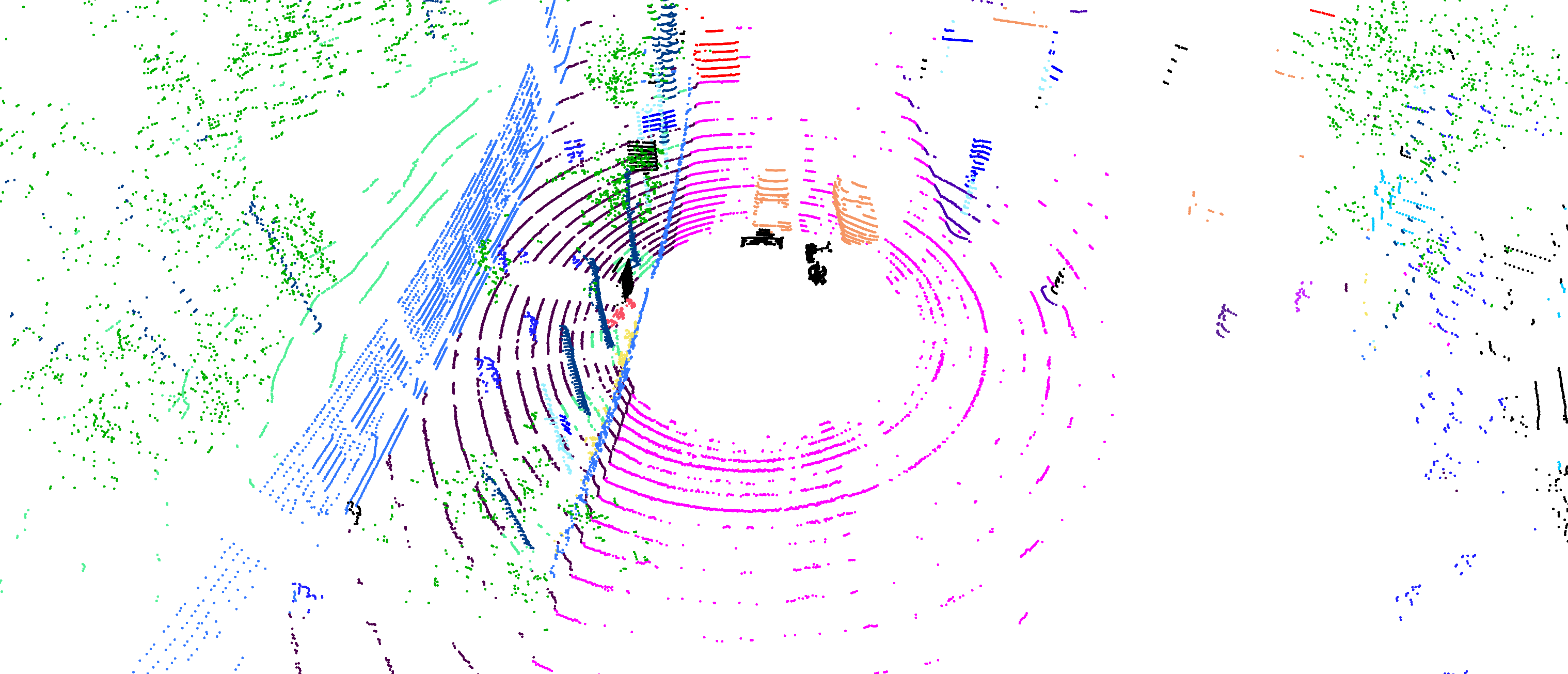}
\hfill
\includegraphics[trim={25cm 5cm 25cm 5cm},clip,width=0.15\textwidth]{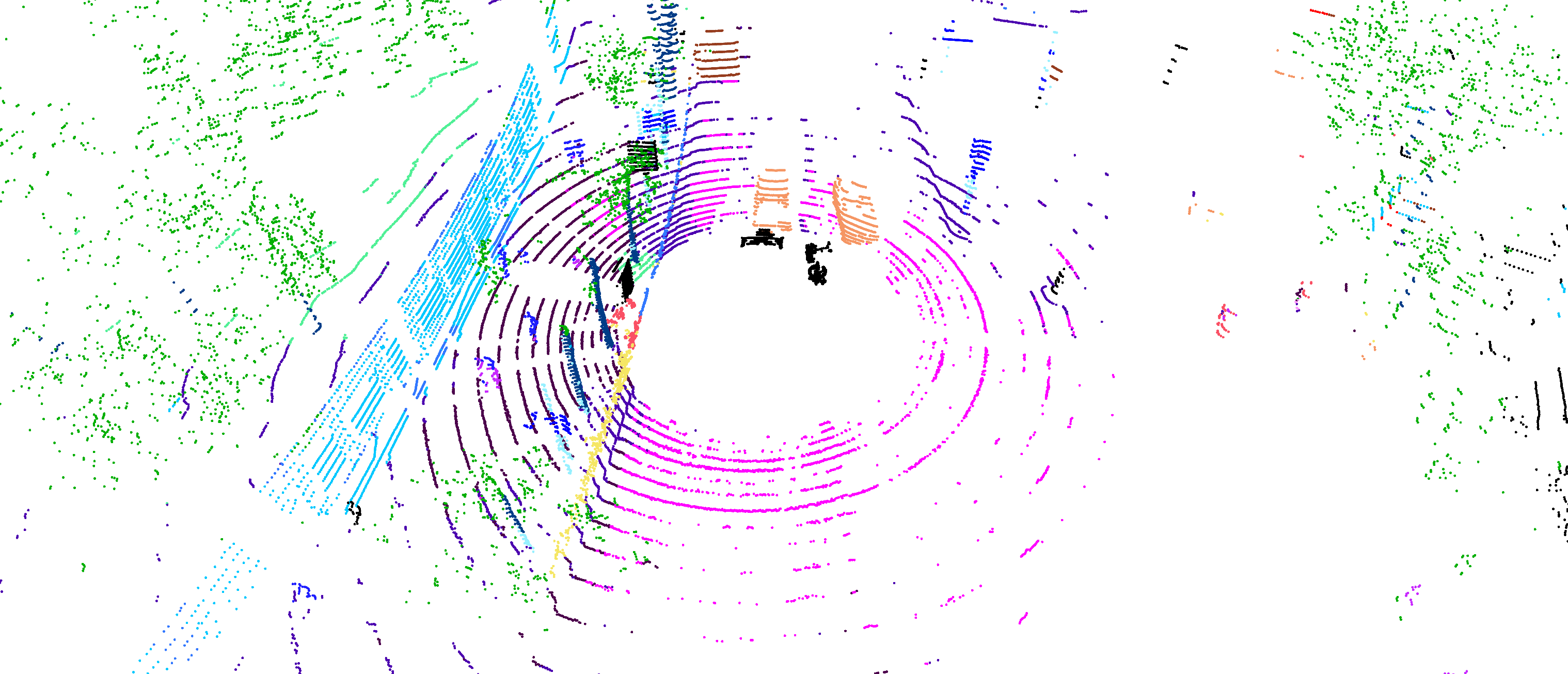}
\hfill
\includegraphics[trim={25cm 5cm 25cm 5cm},clip,width=0.15\textwidth]{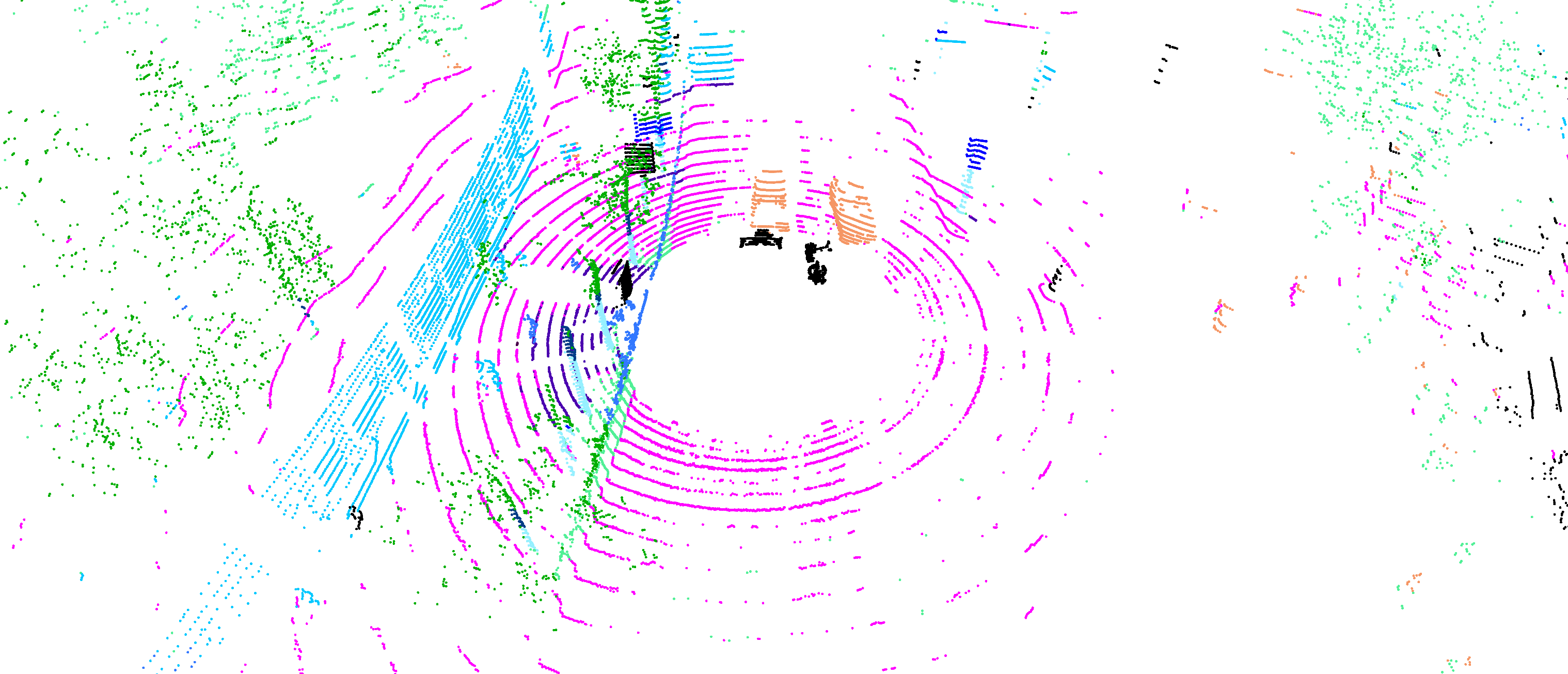}
\hspace{0.01\textwidth}
\textbf{(d)}\hspace{1mm}
\includegraphics[trim={25cm 5cm 25cm 5cm},clip,width=0.15\textwidth]{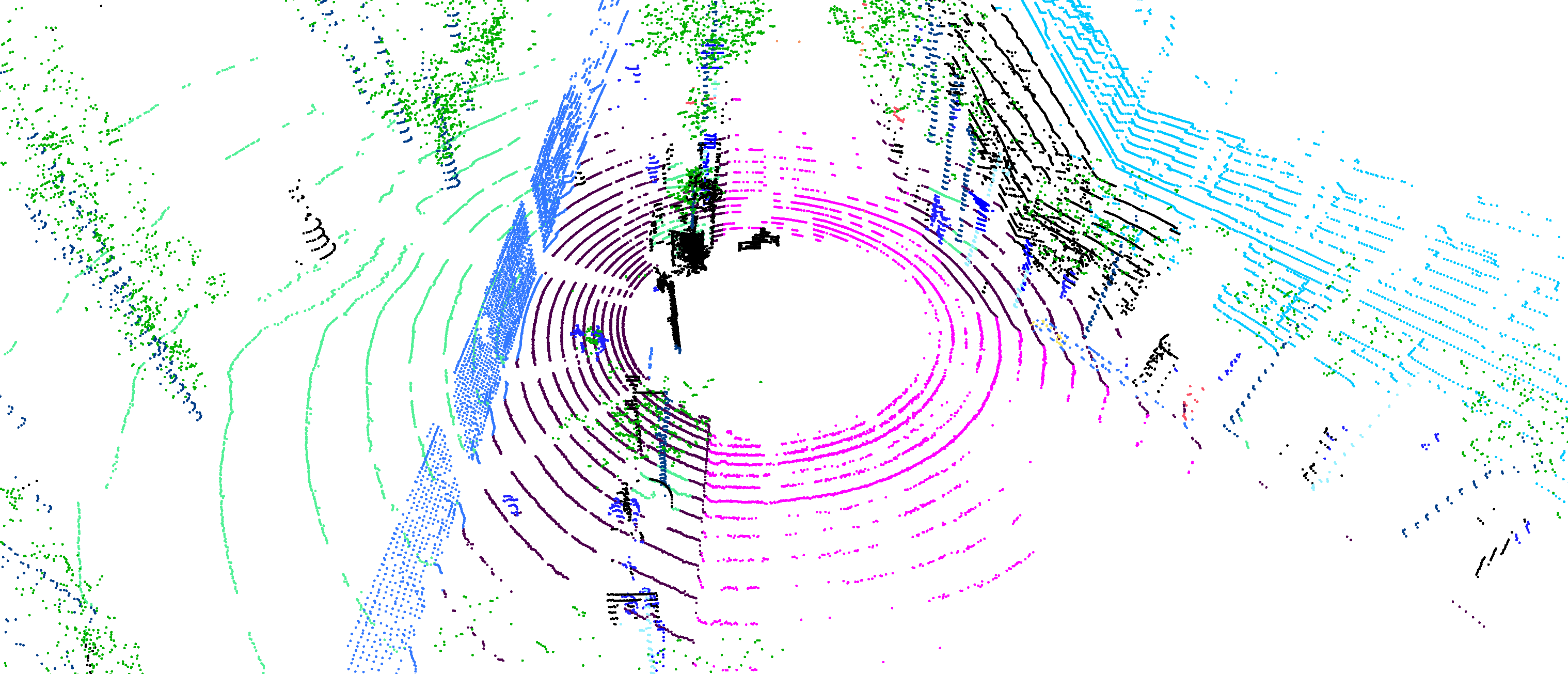}
\hfill
\includegraphics[trim={25cm 5cm 25cm 5cm},clip,width=0.15\textwidth]{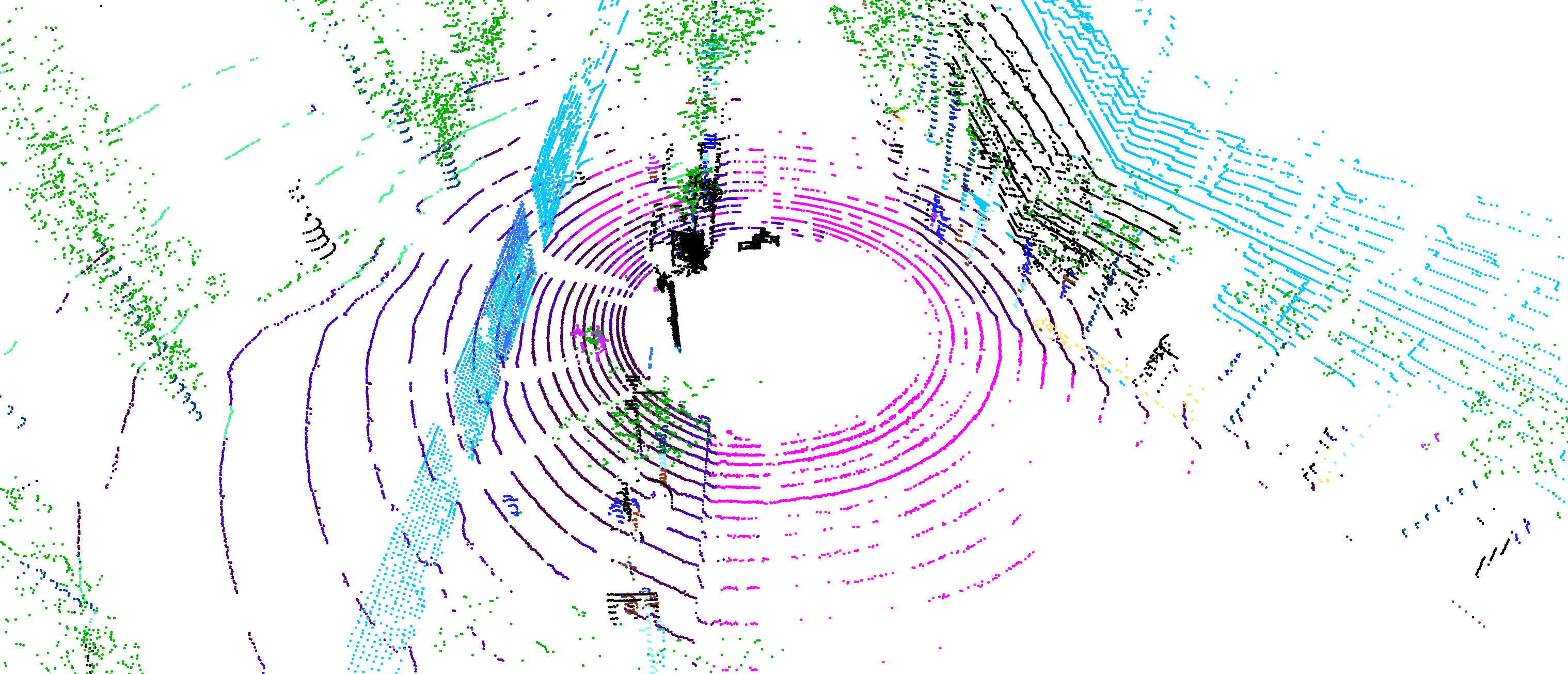}
\hfill
\includegraphics[trim={25cm 5cm 25cm 5cm},clip,width=0.15\textwidth]{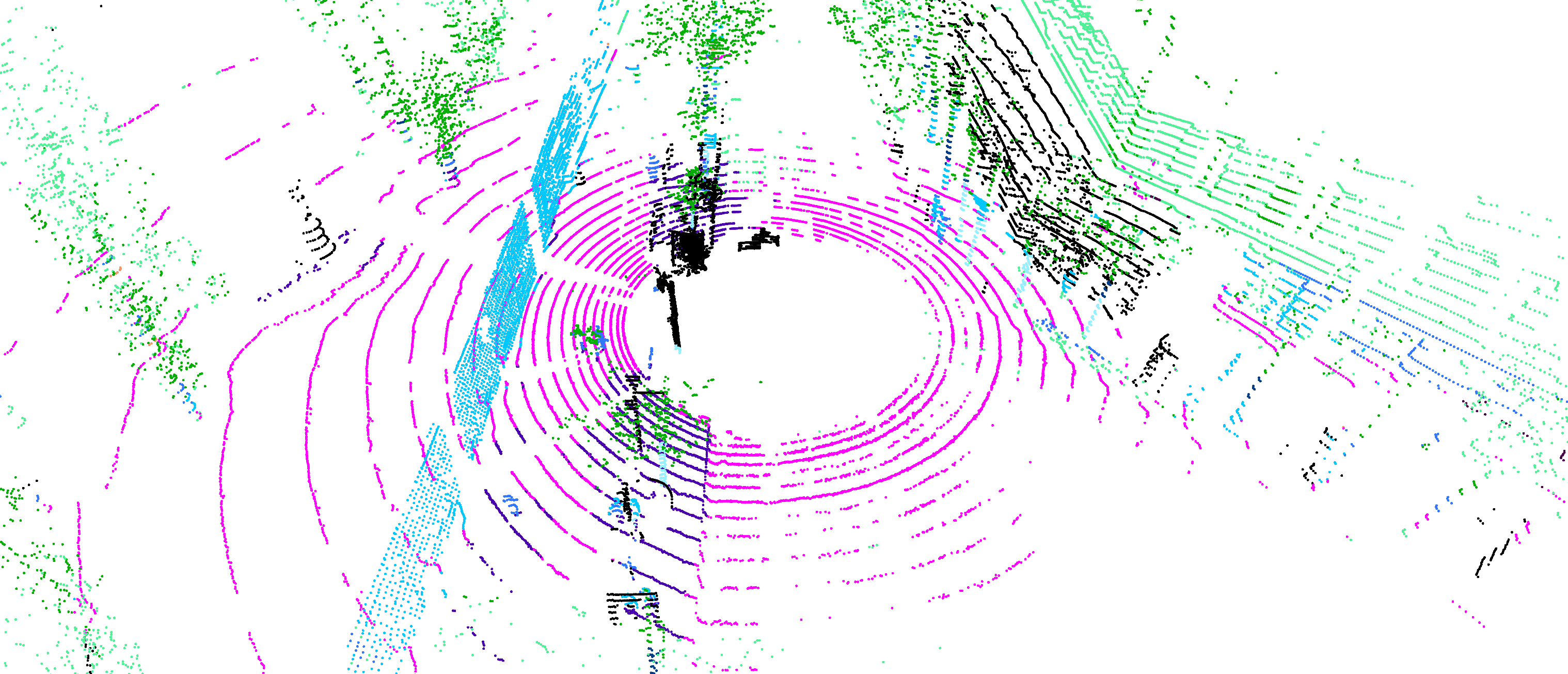}
\end{minipage}

\vspace{1mm}

\caption{
Zero-shot transfer to ParisLuco3D from SemanticKITTI using the same PTv3~\shortcite{ptv3} backbone. DOS shows superior generalization, recovering buses (a, b), reducing wall/building confusion, and separating road from sidewalk. Remaining issues include railing–bike confusion (c) and similar classes swapping.
}
\label{fig:qualitative}
\end{figure*}

To assess generalization to unseen domains, we test on the ParisLuco~\cite{parisluco3d} dataset. In the zero-shot setting, we fine-tune on nuScenes or SemanticKITTI and evaluate directly on ParisLuco. In Table~\ref{tab:parisluco}, we observe that DOS significantly outperforms a supervised-from-scratch baseline using the same labels. In the few-shot case, fine-tuning with only five labeled scenes yields state-of-the-art performance. Figure~\ref{fig:qualitative} compares zero-shot transfer from SemanticKITTI to ParisLuco3D. Despite no adaptation, DOS yields notably improved segmentation over the supervised PTv3 baseline, demonstrating stronger generalization to out-of-distribution urban scenes. To support downstream adoption, we release our multi-dataset pretrained weights as a general-purpose LiDAR backbone for cross-domain transfer.

\begin{table}
    \centering
    \footnotesize
    \setlength\tabcolsep{9.0pt}
    \begin{tabular}{lccc}
        \toprule
        \textbf{Model} & \textbf{mIoU} & \textbf{mACC} & \textbf{ACC}  \\
        \midrule
        Masked self-distillation & & &  \\ 
        \qquad naive & 54.7 & 66.7& 88.9  \\
        \qquad token jitter & 55.1 & 68.3 & 89.2  \\  % ~\cite{sonata} (Sonata)
        \rowcolor[HTML]{E8F5E9} % light green
        \qquad observable (ours)& 69.3& 81.7&93.4  \\  % ~\cite{sonata} (Sonata)
        \midrule
        Training target\\
        \qquad clustering  &69.3& 81.7&93.4  \\ 
        
        \qquad feature regression  &63.0&76.6 &92.2  \\ 
        \rowcolor[HTML]{E8F5E9} % light green
        \qquad softmaps (ours) & 72.3 & 83.4 &94.0  \\ 
        \midrule
        + Zipf prior & & &  \\ 
        \qquad uniform (alpha = 0) &72.3 & 83.4 &94.0 \\ 
        \rowcolor[HTML]{E8F5E9} % light green
        \qquad alpha = 1.3 & \textbf{74.1} & \textbf{84.8}& \textbf{93.7} \\ 
        \bottomrule
    \end{tabular}
    \caption{Ablation on DOS components, evaluated via LP on the nuScenes val set (semantic segmentation).}
    \label{tab:dos_incremental}
\end{table}

\begin{table*}[h]
    \begin{minipage}{.88\columnwidth}
    \footnotesize
    \centering
    \bgroup
        \setlength\tabcolsep{2pt}
        \begin{tabular}{lccccc}
        \toprule
        \textbf{Method} & \textbf{0.1\%} & \textbf{1\%} & \textbf{10\%} & \textbf{50\%} & \textbf{100\%} \\
        \midrule
        Supervised & 28.2 & 41.6 & 68.7 & 78.7 & \textcolor{gray}{80.4} \\
        NOMAE \shortcite{sonata}(ft) & 35.8 & 48.1 & 69.9 & 80.1 & \textbf{\textcolor{gray}{81.8}} \\
        \rowcolor[HTML]{E8F5E9}
        DOS (ours, ft) & \textbf{38.8} & \textbf{53.1} & \textbf{71.5} & \textbf{80.5} & \textbf{\textcolor{gray}{81.8}} \\
        \bottomrule
        \end{tabular}
    \egroup
    \caption{Results with varying amounts of annotated data, evaluated for semantic segmentation mIoU on the nuScenes val set.}
    \label{tab:nuscenes_efficient}
    \end{minipage}
    \hspace{.8em}
    \begin{minipage}{1.15\columnwidth}
    \footnotesize
    \centering
    \bgroup
    \setlength\tabcolsep{2pt}
    \begin{tabular}{lccccccccccc}\toprule
       \textbf{Method} &\multicolumn{5}{c}{\textbf{Limited Scenes}} &\multicolumn{5}{c}{\textbf{Limited Annotations (pts)}}\\\cmidrule(lr){2-6}
       \cmidrule(lr){7-11}
         & \textbf{1\%} & \textbf{5\%}& \textbf{10\%} & \textbf{20\%} & \textbf{100\%} & \textbf{20}& \textbf{50} & \textbf{100} & \textbf{200} & \textbf{full}\\\midrule
        Supervised &25.8 & 48.9 & 61.0 & 67.0 & \textcolor{gray}{77.2} & 60.1 & 67.9 & 71.4 & 72.7 & \textcolor{gray}{77.2} \\
        Sonata \shortcite{sonata}(lin) & 43.6 & 62.5& 68.6 & 69.8 & \textcolor{gray}{72.5} & 69.0 & 70.5 & 71.1 & 71.5 & \textcolor{gray}{72.5}\\
        \rowcolor[HTML]{E8F5E9} % light green
        DOS (ours, lin) & \textbf{45.8} & \textbf{63.4} & \textbf{69.7} & \textbf{70.9} & \textbf{\textcolor{gray}{72.8}} & \textbf{69.4} & \textbf{70.8} & \textbf{71.4} & \textbf{71.7} & \textbf{\textcolor{gray}{72.8}} \\
        \bottomrule
    \end{tabular}
    \egroup
    \caption{Results with varying amounts of annotated data, evaluated for semantic segmentation mIoU on the ScanNet val set.}
    \label{tab:data_efficient}
    \end{minipage}
\end{table*}

\section{Ablation Study}

We conduct ablation studies to evaluate the design choices behind DOS and justify each component’s contribution. In addition to our default PTv3 backbone, we assess DOS with an alternative architecture to test generality. Further analysis on centering strategies, label-efficient settings, and prototype count is presented in the supplementary material.

\subsubsection{Evaluating Components of DOS.}
\label{sec:ingredients_of_dos}

We evaluate the contribution of each component in DOS via an incremental ablation study on the nuScenes dataset, using LP for semantic segmentation after 50 epochs of pretraining. All models use a PTv3 backbone with consistent hyperparameters, including a 70\% masking ratio and fixed block size.
We begin with a standard \textit{masked self-distillation} baseline that supervises the student via an online clustering loss. This naive setup yields 54.7 mIoU and suffers from positional leakage through masked tokens, as also observed by prior works~\cite{sonata,msm}. Token jittering~\cite{sonata}, proposed as a mitigation, improves this only slightly to 55.1 mIoU.
Replacing this with our proposed \textit{observable self-distillation}, which discards masked tokens and supervises only visible voxels, leads to a substantial improvement of +14.6 mIoU, achieving 69.3 mIoU. Unlike Sonata, our approach does not rely on auxiliary heads or additional labeled data, yet achieves significantly higher performance under this restricted distillation setup.

Next, we assess the impact of the distillation target. Keeping the observable setup fixed, we replace clustering with a \textit{feature regression} loss (e.g., cosine similarity), which degrades performance to 63.0 mIoU. By contrast, our proposed \textit{softmap distillation} boosts results to 72.3 mIoU, a +3.0 improvement over clustering, demonstrating the benefits of spatially-aware supervision and inter-point competition.
Finally, we integrate a Zipfian prior over prototype usage via \textit{Zipf-Sinkhorn}, yielding a further improvement from 72.3 to 74.1 mIoU. This suggests that aligning prototype assignments with a Zipfian prior, by better capturing the long-tail distribution of 3D semantics, enhances semantic diversity and representation quality. We adopt this final setup in all main experiments.

\begin{table}
\centering
\footnotesize
\bgroup
\setlength\tabcolsep{1pt}
    \begin{tabular}{l|c|c|c|c|c|c|c|c|c}
    \toprule
    \textbf{Zipf Exponent \(\alpha\)} & \textbf{0.0} & \textbf{0.1} & \textbf{0.3} & \textbf{0.6} & \textbf{0.9} & \textbf{1.3} & \textbf{1.6} & \textbf{2.0} & \textbf{3.0} \\
    \midrule
    ScanNet (mIoU)    & 71.9 & 72.2 & 71.9 & 72.3 & 72.5 & \textbf{72.8} & 72.4 & 71.9 & 69.8 \\
    ScanNet200 (mIoU) & 27.9 & 28.1 & 28.2 & 28.6 & 28.7 & 29.1 & \textbf{29.2} & 29.0 & 27.7 \\
    \bottomrule
    \end{tabular}
\egroup
\caption{Ablation of Zipf exponent \(\alpha\) on ScanNet and ScanNet200. Reported scores are mIoU under LP.}
\label{tab:zipf_ablation}
\end{table}
\begin{table}
\centering
\footnotesize
\bgroup
\setlength\tabcolsep{13.5pt}
\begin{tabular}{lccc}
\toprule
\(\alpha\) & Head (66) & Common (68) & Tail (66) \\
\midrule
0.0 & 50.4 & 20.5 & 10.6 \\
1.3 & 50.8 & 23.5 & 13.2 \\
\bottomrule
\end{tabular}
\egroup
\caption{Effect of Zipf exponent \(\alpha\) on ScanNet200 head/common/tail splits (66/68/66 classes). Reported scores are mIoU under LP.}
\label{tab:zipf_headtail}
\end{table}

\subsubsection{Label-Efficiency}

We evaluate DOS under limited supervision settings to assess its robustness in data-scarce regimes. Table~\ref{tab:nuscenes_efficient} and Table~\ref{tab:data_efficient} report semantic segmentation results on nuScenes and ScanNet across various label budgets.
On both benchmarks, DOS consistently outperforms supervised training from scratch and prior SSL methods such as Sonata and NOMAE. In the linear probing setting, DOS achieves strong performance even with only 1\% of labeled scenes, and reaches parity with full supervision at just 20\% labeling. In the fine-tuning setting, DOS-pretrained models yield the highest accuracy across all label fractions, including extremely low-resource cases (e.g., 0.1\% on nuScenes).
These results demonstrate that DOS learns semantically rich and transferable representations that enable effective downstream adaptation with minimal annotation.

\subsubsection{Effect of Zipf Exponent.}

We evaluate the impact of the Zipf exponent \(\alpha\) in Zipf-Sinkhorn regularization on ScanNet and ScanNet200, which share geometry but differ in label granularity (20 vs. 200 classes). This setup helps us assess how prior sharpness influences representation quality across coarse and fine semantic spaces.
As shown in Table~\ref{tab:zipf_ablation}, moderate values of \(\alpha\) yield the best performance: mIoU peaks at \(\alpha = 1.3\) on ScanNet (72.8) and \(\alpha = 1.6\) on ScanNet200 (29.2). Lower \(\alpha\) values (\(\approx 0\)) enforce overly uniform usage, while higher values (\(\geq 2\)) degrade performance and cause training instability due to prototype under-utilization and weakened gradients.
To better understand where the gains on ScanNet200 come from, we further break down performance using the standard head/common/tail split of 66/68/66 classes based on class frequency (Table~\ref{tab:zipf_headtail}). When increasing \(\alpha\) from 0.0 to 1.3, performance on head classes remains nearly unchanged (50.4~\(\rightarrow\)~50.8 mIoU), whereas common and tail classes improve substantially (20.5~\(\rightarrow\)~23.5 and 10.6~\(\rightarrow\)~13.2 mIoU, respectively). This indicates that mild Zipf priors mainly help by strengthening representations for medium-frequency and rare categories rather than further boosting already well-represented head classes.
These results align with empirical Zipfian statistics observed in real-world 3D datasets, and support the use of mild Zipf priors to balance semantic coverage, specialization, and stability.\looseness=-1

\subsubsection{Effect of Prototype Count.}

We study how the number of prototypes affects representation quality for three objectives: clustering, softmap distillation, and softmap with Zipf-Sinkhorn. As shown in Table~\ref{tab:prototype_count_ablation}, softmap-based objectives consistently outperform clustering across all settings, with clear gains even at low prototype counts. This highlights the benefit of modeling contextual relevance and inter-point competition in softmaps compared to the point-wise, independent clustering assignments. Softmap variants also saturate early, typically between 1024 and 4096 prototypes, indicating better training efficiency and stability. In contrast, clustering keeps improving with more prototypes but remains worse overall, and further gains would likely require impractically large cluster counts due to memory constraints.

\begin{table}[h]
\centering
\footnotesize
\begin{tabular}{lcccc}
\toprule
\textbf{Prototype Count} & \textbf{Clustering} & \textbf{Softmap} & \textbf{Softmap + Zipf} \\
\midrule
\multicolumn{4}{l}{\textit{nuScenes}} \\
32     & 65.1 & 70.9 & 70.2 \\
128    & 69.5 & 69.8 & 73.3 \\
\rowcolor[HTML]{E8F5E9} % light green
1024   & 68.7 & 72.1 & \textbf{74.1} \\
4096   & 69.6 & 72.3 & 74.0 \\
\midrule
\multicolumn{4}{l}{\textit{ScanNet}} \\
32     & 59.6 & 66.1 & 65.0 \\
128    & 62.7 & 70.8 & 70.6 \\
\rowcolor[HTML]{E8F5E9} % light green
1024   & 66.7 & 71.5 & 72.8 \\
4096   & 69.3 & 71.7 & \textbf{72.9} \\
\bottomrule
\end{tabular}
\caption{LP mIoU (\%) on nuScenes and ScanNet for varying prototype counts and objectives.}
\label{tab:prototype_count_ablation}
\end{table}

\begin{table}[h]
\centering
\footnotesize
\begin{tabular*}{\columnwidth}{@{\extracolsep{\fill}}lcc}
\toprule
\textbf{Dataset} & \textbf{w/o Cross-View} & \textbf{Full DOS (Ours)} \\
\midrule
nuScenes & \textbf{74.3} & 74.1 \\
Waymo    & 63.5 & \textbf{66.1} \\
ScanNet  & 61.8 & \textbf{72.8} \\
\bottomrule
\end{tabular*}
\caption{Effect of cross-view supervision on linear probing performance (mIoU \%) across datasets.}
\label{tab:cross_view_ablation}
\end{table}

\subsubsection{Effect of Cross-View Supervision.}

We ablate cross-view supervision in DOS by training a variant without the view alignment objective. For a fair comparison, we double the batch size and number of training epochs so that the total compute matches the default setup, and evaluate on nuScenes, Waymo, and ScanNet with linear probing. As shown in Table~\ref{tab:cross_view_ablation}, the effect of cross-view supervision is dataset dependent: on ScanNet, with dense indoor point clouds, removing cross-view alignment causes a substantial performance drop, while on sparser outdoor datasets such as nuScenes and Waymo, the effect is minimal or slightly negative. This suggests that cross-view supervision is most beneficial when point cloud density supports meaningful spatial alignment across views.

\subsubsection{Transferability Across Architectures.}

We evaluate DOS on SPUNet to test generality beyond PTv3, using the same pretraining and evaluation. As shown in Table~\ref{tab:architecture_transfer}, DOS significantly improves SPUNet finetuning over training from scratch, confirming its value as a general purpose initialization. However, LP yields weaker representations than PTv3, suggesting sensitivity to backbone capacity.To investigate further, we double SPUNet channel width, which improves LP scores but still falls short of PTv3. When PTv3 is unavailable or too heavy, we recommend distilling from a pretrained DOS encoder. With just 10 epochs of distillation from a frozen DOS model, SPUNet outperforms its self distilled version, which highlights an efficient path to lightweight 3D backbones.

\begin{table}
\centering
\footnotesize
    \begin{minipage}{0.48\linewidth}
    \centering
    \textbf{Fine-tuning} \\
    \begin{tabular}{lc}
    \toprule
    Method & mIoU \\
    \midrule
    scratch       & 73.3 \\
    UniPAD        & 79.4 \\
    NOMAE         & 80.1 \\
    \rowcolor[HTML]{E8F5E9} % light green
    DOS           & \textbf{80.2} \\
    PTv3 + DOS (ref.)    & \textcolor{gray}{81.8} \\
    \bottomrule
    \end{tabular}
    \end{minipage}
    \hfill
    \begin{minipage}{0.48\linewidth}
    \centering
    \textbf{Linear Probing} \\
    \begin{tabular}{lc}
    \toprule
    SPUNet & mIoU \\
    \midrule
    DOS                 & 55.6 \\
    + double channels   & 57.4 \\
    \rowcolor[HTML]{E8F5E9} % light green
    + distill from PTv3 & \textbf{71.5} \\
    PTv3 + DOS (ref.)   & \textcolor{gray}{74.8} \\
    \bottomrule
    \end{tabular}
    \end{minipage}
\caption{Transferability of DOS across backbone architectures . We report mIoU on nuScenes under LP and FT.}
\label{tab:architecture_transfer}
\end{table}

\section{Conclusion}

We introduced \textit{DOS}, an SSL framework for 3D point clouds that addresses key limitations of existing methods. DOS builds on three core innovations: (1)~\textit{observable self-distillation}, which avoids positional leakage by supervising only visible points, (2)~\textit{semantic softmaps}, which enhance representation learning through point-wise competition and relevance modeling, and (3)~\textit{Zipf-Sinkhorn}, a prototype assignment strategy aligned with the naturally long-tailed distribution of real-world semantics. Together, these components enable DOS to learn high-quality 3D features without labels. DOS outperforms SOTA across five challenging 3D benchmarks and shows strong generalization in cross-domain and few-shot settings. We release a general-purpose LiDAR backbone pretrained across multiple datasets to accelerate progress in 3D scene understanding.

\textbf{Limitations and Future Work.} While DOS currently handles indoor and outdoor domains separately, unifying them within a single pretraining framework remains a promising direction. Moreover, although DOS closely approaches supervised performance under LP, surpassing this baseline is a challenge for future research in 3D representation learning.

\bibliography{aaai2026}

\end{document}